\documentclass{article}

\PassOptionsToPackage{numbers}{natbib}

    \usepackage[preprint]{neurips_2020}

\usepackage[utf8]{inputenc} 
\usepackage[T1]{fontenc}    
\usepackage{hyperref}       
\usepackage{url}            
\usepackage{booktabs}       
\usepackage{amsfonts}       
\usepackage{nicefrac}       
\usepackage{microtype}      

\usepackage{xspace}
\newcommand{\system}[0]{\textsc{NetFuse}\xspace}

\usepackage{graphicx}
\usepackage{subcaption}
\usepackage{wrapfig}
\usepackage{multirow}
\usepackage{amsmath}
\usepackage{amssymb}
\usepackage{amsthm}
\usepackage{bm}
\usepackage{mathtools}
\usepackage{algorithm}
\usepackage{algorithmic}

\DeclarePairedDelimiter\floor{\lfloor}{\rfloor}
\newcommand\numberthis{\addtocounter{equation}{1}\tag{\theequation}}

\usepackage{lipsum}

\usepackage{xcolor}

\title{Accelerating Multi-Model Inference by \\ Merging DNNs of Different Weights}

\author{%
  Joo Seong Jeong \\
  Seoul National University \\
  \And
  Soojeong Kim \\
  Seoul National University \\
  \AND
  Gyeong-In Yu \\
  Seoul National University \\
  \And
  Yunseong Lee\thanks{Yunseong is currently affiliated with Qualcomm AI Research.} \\
  Seoul National University \\
  \And
  Byung-Gon Chun \\
  Seoul National University \\
}

\begin{document}

\maketitle

\begin{abstract}
Standardized DNN models that have been proved to perform well on machine learning tasks are widely used and often adopted as-is to solve downstream tasks, forming the transfer learning paradigm.
However, when serving multiple instances of such DNN models from a cluster of GPU servers, existing techniques to improve GPU utilization such as batching are inapplicable because models often do not share weights due to fine-tuning.
We propose \system, a technique of merging multiple DNN models that share the same architecture but have different weights and different inputs.
\system is made possible by replacing operations with more general counterparts that allow a set of weights to be associated with only a certain set of inputs.
Experiments on ResNet-50, ResNeXt-50, BERT, and XLNet show that \system can speed up DNN inference time up to 3.6$\times$ on a NVIDIA V100 GPU, and up to 3.0$\times$ on a TITAN Xp GPU when merging 32 model instances, while only using up a small additional amount of GPU memory.
\end{abstract}
\section{Introduction}

Various standardized deep neural network (DNN) models exist for modern machine learning tasks.
For example, attention models such as BERT and XLNet have recently been proven to be particularly effective for language understanding tasks~\citep{bert, xlnet}.
Meanwhile, the ResNet, Inception, and ResNeXt models are widely used for image classification tasks~\citep{resnet, inception, resnext}.
Such models are recognized for their ability to learn general representations for their respective input data distributions, and are often used to solve new tasks with little to no modifications in model architecture.
The applicability of these representative models is backed by the \emph{transfer learning} paradigm -- knowledge earned by a model while training on one task can be passed down to another model training on some other related task.
In fact, pretrained parameters for these models (on major datasets) are publicly available online in the form of model zoos.

Before being deployed to serving systems, such models undergo a \emph{fine-tuning} process in which the parameters of the models are altered specifically for the task in hand.
This process is necessary because the model output required by a specific task often does not align with that of the general task employed to learn the pretrained parameters.
For example, a ResNet~\citep{resnet} trained on a 1000-class dataset~\citep{imagenet} cannot be adopted as-is for a 10-class dataset~\citep{cifar10}; the last classification layer must be replaced with an appropriate substitute, followed by additional training.
Fine-tuning also brings the benefit of pushing the model parameters towards the specific task, making the model lose generality but gain specialty.
This results in a situation where many specialized models have similar architectures but different internal parameters.

Many systems for serving DNN models on CPUs and GPUs have been proposed, each focusing on a different aspect of DNN serving, or DNN \emph{inference}.
When it comes to serving multiple models from a cluster of servers, DNN inference is known to have poor resource utilization, especially for GPUs~\citep{jain2018dynamic}.
This is mainly due to how the computation capabilities of modern GPUs are usually more than enough to handle the actual amount of computation required during inference.
Unlike training, DNN inference does not involve the usual ``backward pass'' in which gradients are calculated from a certain loss function, resulting in a much less amount of floating point operations (FLOPs).

A common technique to improve resource utilization during DNN inference is \emph{batching}~\citep{clipper, pretzel, mcdnn}.
Thanks to the sheer amount of cores within a GPU, simply providing the GPU with many, mutually independent operations at once by batching inputs is actually a good approach to boost GPU utilization.
However, conventional batching is confined to only single-model settings, limiting the applicability of batching in the context of serving systems that deal with multiple models.
There have been recent systems that propose multi-model batching~\citep{mainstream, nexus} as an alternative, but this is not feasible when models have completely different parameters, which is in fact a relatively common case considering transfer learning and fine-tuning.
Without batching (both single-model and multi-model), models must be run in isolation from one another, missing out on optimization opportunities.

In this paper, we propose \system, a technique of coalescing weights to merge multiple DNN models of different parameters\footnote{The terms \emph{weight} and \emph{parameter} are interchangeable within the context of DNNs. We use both terms throughout this paper.}
as well as different inputs.
Our technique fuses several instances of a DNN operation, that do not share weights, into one larger operation by carefully aligning the weights so that a set of weights is only associated with its original corresponding input.
Since the act of associating a set of weights with only a certain input is not allowed for some types of operations, we substitute such operations with more general counterparts, if necessary -- e.g., replace layer normalization with group normalization, and convolution with grouped convolution.
Unlike batching, \system does not require models to share parameters or inputs.

We evaluate \system using four representative models (ResNet-50, ResNeXt-50, BERT, XLNet) for vision and language tasks on two types of NVIDIA GPUs (V100, TITAN Xp).
Under various settings that merge different number of models with different batch sizes, \system outperforms other baselines by up to 3.6$\times$ in terms of inference time.

\section{Related Work} \label{sec:back}

\subsection{Multi-Model Inference}

\begin{figure}[t]
  \centering
  \includegraphics[width=0.5\textwidth]{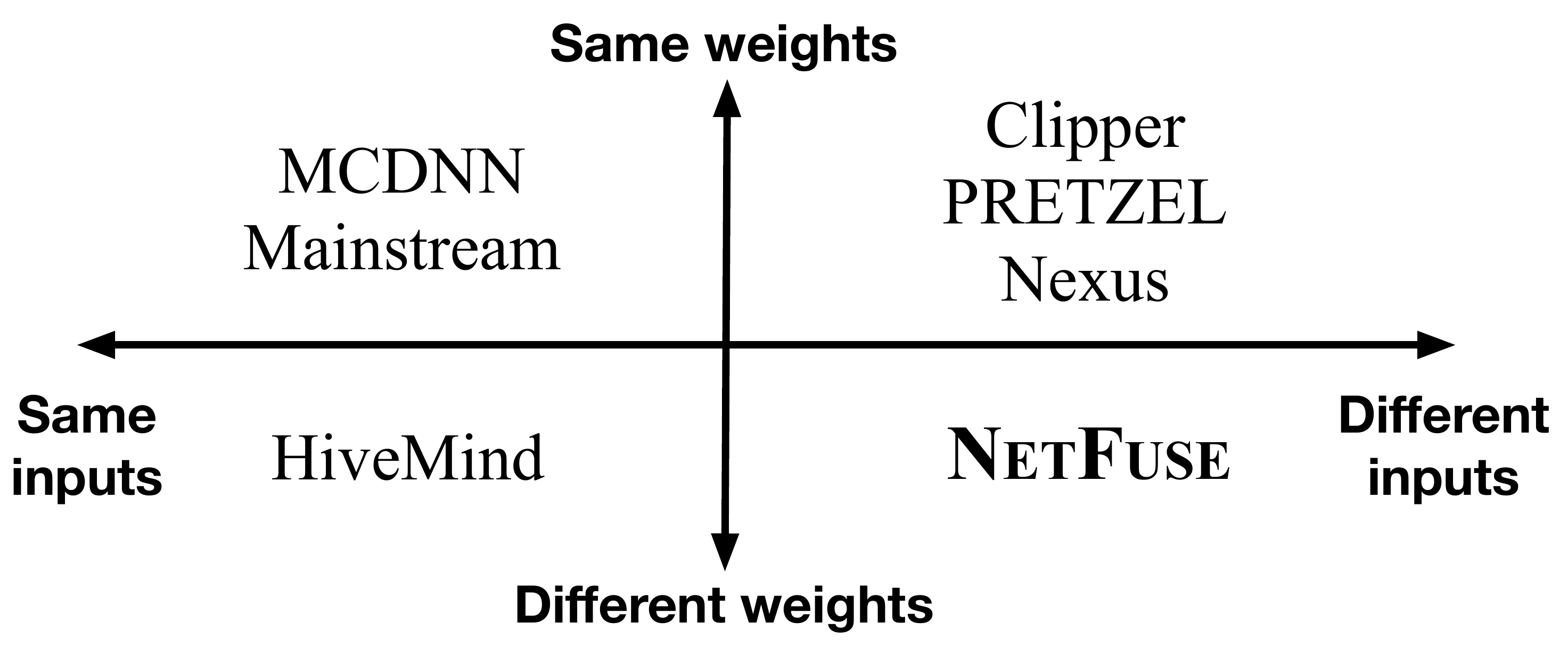}
  \caption{Comparison of serving systems that employ computation merging. \system is the only one that is able to merge computations when both inputs and weights are different.}
  \label{fig:systems}
\end{figure} 

Many inference systems have proposed solutions for improving the GPU utilization of DNN inference by either reusing computations when possible, or merging several computations into one large computation~\citep{clipper, pretzel, mcdnn, mainstream, nexus, hivemind}.
Different systems assume different input and model settings, leading to different merging scenarios and strategies (Figure~\ref{fig:systems}).

\textbf{Same inputs, same weights.}
The inference result for a certain query can be reused as-is for another query if both the query input and the target DNN weights are identical.
Referred to as \emph{model sharing} by MCDNN~\citep{mcdnn} and Mainstream~\citep{mainstream}, this technique can be applied to cases where a single input is processed by multiple DNNs of similar tasks, e.g., age classification and gender classification on human faces.
The system identifies a common subnetwork from multiple DNNs and runs the input through the common subnetwork only once so that the same computation is not repeated needlessly.

\textbf{Different inputs, same weights.}
Inference queries on different inputs can also be merged if the weights of the target DNN are the same, in the form of \emph{batching}.
Batching several DNN inputs together is a classic technique for exploiting the GPU's parallel computing power, and many DNN operations are implemented with batch size in mind.
Systems such as Clipper~\citep{clipper} and PRETZEL~\citep{pretzel} employ batching by delaying the processing of a certain query and merging the query with subsequent queries, improving inference throughput with the cost of sacrificing latency.
Another system, Nexus~\citep{nexus}, demonstrates a similar batching technique of aggregating inputs from multiple queries on different DNNs, assuming the DNNs have large subnetworks in common.

\textbf{Same inputs, different weights.}
HiveMind~\citep{hivemind} introduces \emph{cross-model layer fusion}, a technique of merging DNNs of different weights.
Instead of batching inputs, HiveMind batches the weights and applies them to the same input.
In essence, cross-model layer fusion can be regarded as the weight-input counterpart of batching.

\textbf{Different inputs, different weights.}
All aforementioned systems require one major condition: when merging computations, either the inputs or the DNN weights (or both) must be identical.
This condition prohibits merging computations of different inputs and different weights.

Consider the following situation: a server is serving instances of the BERT~\citep{bert} model for several natural language processing tasks such as question answering, sentence prediction, and text generation.
Each task demands its own \emph{fine-tuning} procedure, i.e., the DNN weights of each task are different.
Moreover, given the nature of the tasks, each task is associated with a different input stream.
Existing approaches on computation merging are inapplicable here, because of the input/weight differences.

Researchers have demonstrated various benefits of fine-tuning all DNN weights in many real-world use cases.
For example, in natural language processing, the state-of-the-art neural networks fine-tune all parameters of a pre-trained model for each downstream task~\citep{bert, radford2018improving, xlnet, howard2018fine}.
Fine-tuning is also shown to be effective in computer vision~\citep{yosinski2014how, chu2016best, zhou2017fine}.
Classic model ensemble techniques~\citep{lee2015m,lakshminarayanan2017simple} utilize models of the same architecture and different weights as well;
at test time, the inference results of each model are aggregated to produce the final result.
Some researches have even proposed branched models~\citep{spottune,hydranet} which contain specialized subnetworks of the same architecture.
Previous computation merging techniques are invalid for all such cases, despite the DNN architecture being largely unchanged.

\begin{wrapfigure}{r}{0.3\textwidth}
  \centering
  \includegraphics[width=0.3\textwidth]{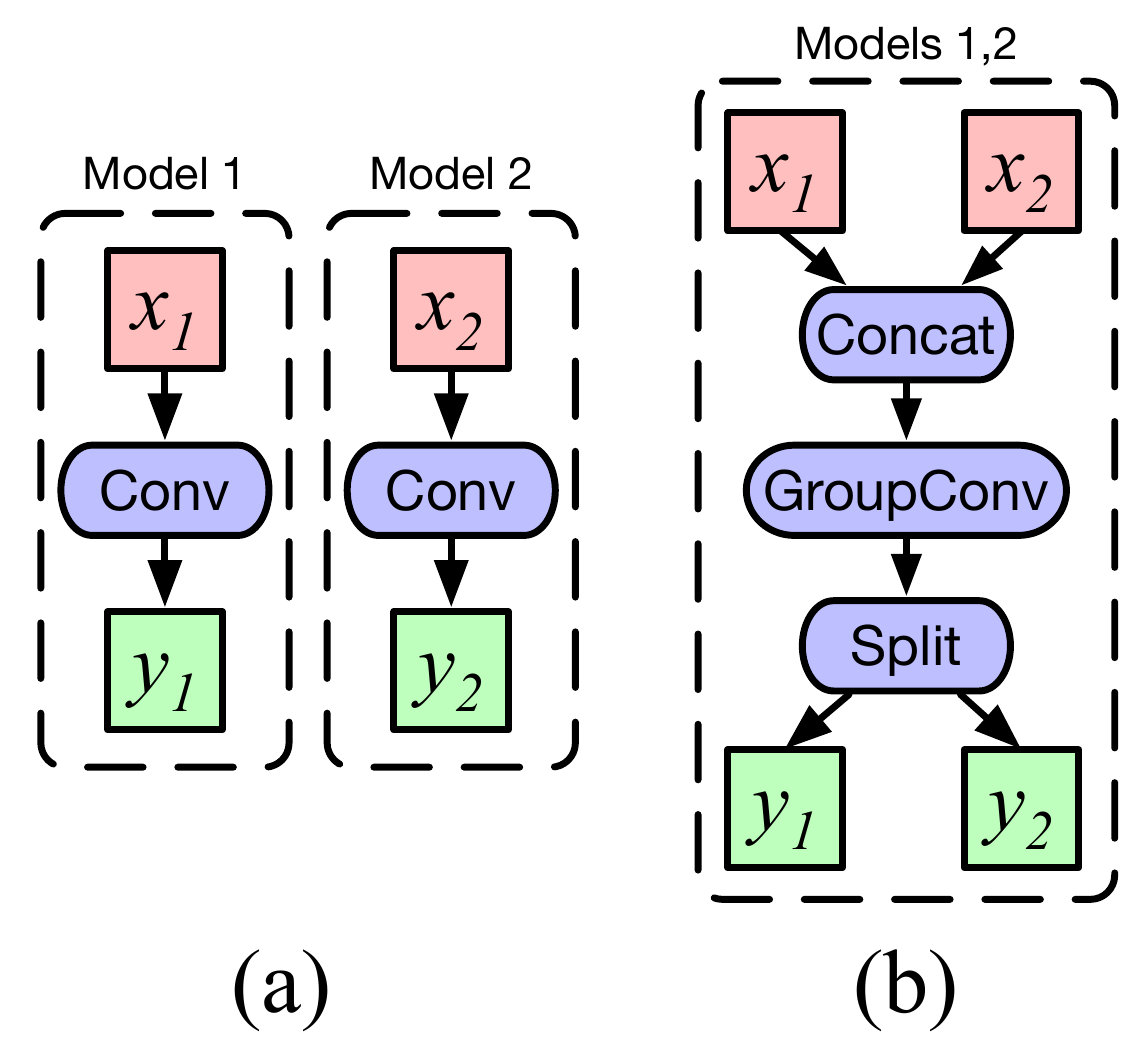}
  \caption{A multi-model optimization example.}
  \label{fig:min-slim}
\end{wrapfigure}

\subsection{Graph Rewriting Frameworks}

A recent line of frameworks -- TASO~\citep{taso}, TVM~\citep{tvm}, TensorFlow XLA~\citep{tf-xla} and TensorRT~\citep{tensorrt} -- propose graph rewriting as a method for optimizing DNN models.
Graph rewriting frameworks apply graph substitution rules, either hand-written or automatically generated, to a DNN model and generate a new model that outputs mathematically equivalent results but can be executed faster on accelerators.
At first glance, graph rewriting seems like a good approach for optimizing the aforementioned use case of models with different inputs and weights.
Since a set of $M$ disjoint models, with one input each, can be considered as one large model with $M$ inputs, we can simply feed the models as-is into a graph rewriting framework for optimization and hope the framework cleverly merges operations across models.

Unfortunately, we find that existing graph rewriting frameworks are ineffective for supporting multi-model inference.
First, the greedy search strategies of existing frameworks prefer single-model optimizations over multi-model optimizations, as multi-model optimizations are often hidden behind overheads.
Figure~\ref{fig:min-slim} depicts an instance in which the execution of two convolutions from two separate models in (a) can be accelerated by concatenating the inputs/outputs and merging the convolutions into a grouped convolution, as shown in (b).
TASO~\citep{taso}, a state-of-the-art graph rewriting framework, is unable to discover this optimization, despite the grouped convolution being faster.
We can manipulate TASO into finding this optimization by giving it extremely aggressive search hyperparameters, but this leads us to the next problem.

Second, frameworks that automatically generate and apply substitution rules experience scalability issues when optimizing multiple disjoint models.
For example, TASO takes more than 30 hours to fully explore the optimization space (even under conservative search hyperparameters) when given four instances of ResNeXt-50~\citep{resnext}, and flat-out runs out of memory when given eight instances.
Even then, we found that TASO does not apply any kind of significant multi-model optimization, because TASO's default substitution rules do not cover the multi-model inference case on hand.

\section{\system} \label{sec:tech}

The key to merging multiple instances of an operation with different inputs and different weights, is to find a more general counterpart of the operation that allows a set of weights to be paired with only a certain set of inputs.
Throughout this section, we will use the term \emph{input-weight pair} to indicate a tuple of inputs and weights that are used together, exclusively.
We will also use the term \emph{input-weight local computation} when referring to the computation of a specific input-weight pair.

\begin{figure}[t]
\begin{center}
  \begin{subfigure}[b]{.5\linewidth}
  \centering
  \includegraphics[width=.9\linewidth]{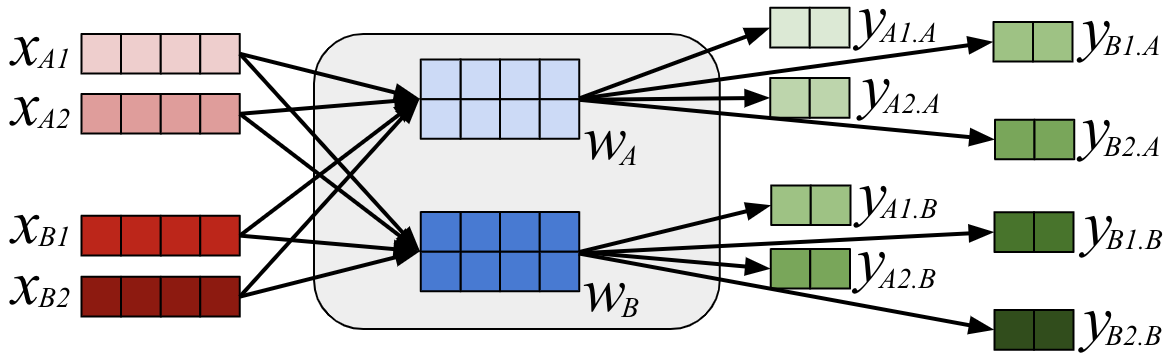}
  \caption{Operation with no local computations}
  \label{fig:ops-global}
  \end{subfigure}%
  \begin{subfigure}[b]{.5\linewidth}
  \includegraphics[width=.9\linewidth]{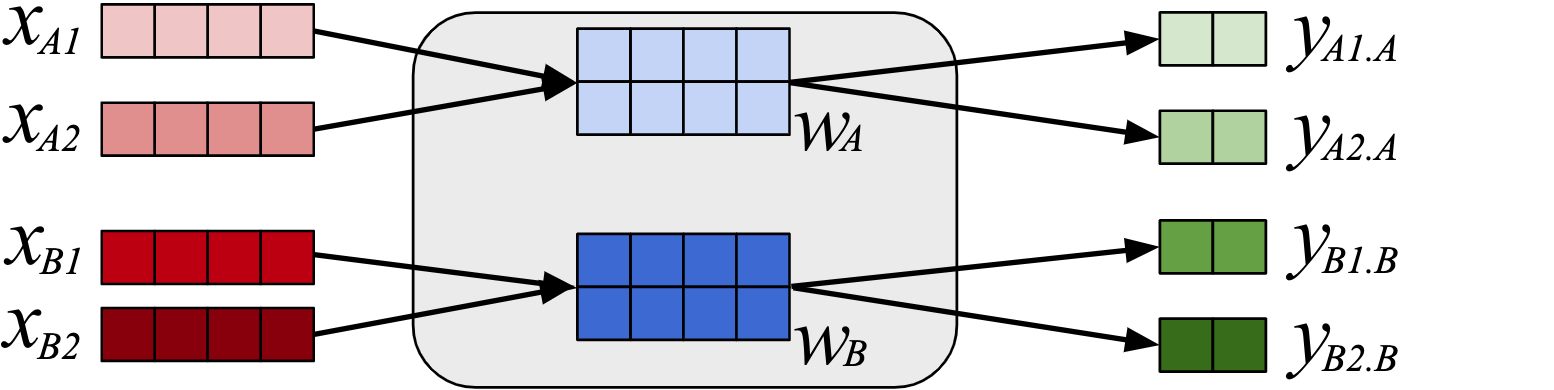}
  \centering
  \caption{Operation with input-weight local computations}
  \label{fig:ops-local}
  \end{subfigure}

  \caption{Structure of operations regarding the pairing of inputs and weights}
  \label{fig:ops}
\end{center}
\end{figure}

We highlight in detail why operations that incorporate input-weight local computations are necessary when merging operations, with Figure~\ref{fig:ops}.
Figure~\ref{fig:ops-global} depicts an abstract illustration of DNN operations whose inputs are associated with all available weights, symbolizing operations such as matrix multiplication (fully connected layers) and convolution.
In the figure, we are trying to merge an operation of inputs $x_A$ and weights $w_A$ with another operation of inputs $x_B$ and weights $w_B$.
Without parting from the structure shown in Figure~\ref{fig:ops-global}, there is virtually no way of preventing inputs $x_B$ from being processed by $w_A$, because all inputs are associated with all weights.
In order to separate $x_A$ and $w_A$ from $x_B$ and $w_B$, we need another operation that involves input-weight local computations.

\begin{table}[t]  
  \caption{List of various DNN operations and their input-weight local computation counterparts.}
  \label{tbl:op-list}
  \centering
  \begin{tabular}{ll}
  \toprule
  \textbf{No local computations} & \textbf{Allows local computations} \\
  \toprule
  Convolution & Grouped Convolution   \\
  \midrule
  Matrix Multiplication & Batch Matrix Multiplication   \\
  \midrule
  Layer Normalization & Group Normalization   \\
  \midrule
  \multirow{4}{*}{--} & Batch Normalization \\
  & Pooling (Max-pooling, ...) \\
  & Activation Functions (ReLU, ...) \\
  & Element-wise Operations \\
  \bottomrule
  \end{tabular}
\end{table}

Figure~\ref{fig:ops-local} portrays another category of DNN operations that consists of multiple input-weight pairs.
With this structure, it is possible to isolate a set of inputs $x_A$ and weights $w_A$ from the other set of inputs and weights.
In fact, major DNN operations all have some form of counterpart operation that performs the same type of computation as the original operation, but additionally allows a certain degree of local computation among inputs and weights, as shown in Table~\ref{tbl:op-list}.

Such operations were not originally designed to be used for merging operations of different weights.
For instance, one of the first well-known usages of the grouped convolution operation is none other than the acclaimed AlexNet~\citep{alexnet}; the authors describe in their paper that the restriction on convolution channels was actually a compromise they had to make because of the memory limitations of the GPU at that time.
Another use of the operation was MobileNet~\citep{mobilenet}, in which the depthwise convolution (an extreme version of grouped convolution) operation was employed in place of an ordinary convolution to reduce FLOPs and allow deployment on mobile devices, with the cost of sacrificing accuracy.
Interestingly, another DNN, ResNeXt~\citep{resnext}, has recently been proposed to experiment with grouped convolutions to improve model accuracy.
While existing applications of grouped convolution can all be classified as attempts to alter the properties (GPU memory, FLOPs, accuracy) of a single convolution operation, we approach from a different point of view and instead apply the operation as a means of merging multiple operations.

For the rest of this section, we look into merging widely used DNN operations and what general counterpart operations they require (Section~\ref{subsec:tech-op}).
We then elucidate how operation merging can be extended to entire DNNs (Section~\ref{subsec:tech-e2e}).

\subsection{Merging Individual Operations} \label{subsec:tech-op}
\paragraph{Matrix multiplication}
Multiple matrix multiplications, i.e., fully connected layers, can be merged into a \emph{batch matrix multiplication}.
Batch matrix multiplication is simply matrix multiplication with a batch of inputs and a batch of weight tensors.
Each input is multiplied with only one weight tensor, which is exactly how we want to merge operations.
In fact, the kernel implementations of matrix multiplication in frameworks such as TensorFlow~\citep{tensorflow} and PyTorch~\citep{pytorch} support batch matrix multiplication by default.
Merging several instances of matrix multiplication is done by first concatenating the inputs and weights along the batch dimension into one big input batch and one big weight batch, respectively, and then replacing the individual operations with a single batch matrix multiplication operation.

It is noteworthy that since a matrix multiplication can be converted to a mathematically equivalent 1x1 convolution, it is also possible to merge several matrix multiplications as if they were convolutions.
However, we have found that this results in very slow inference speed for even moderately sized DNNs, due to how the implementation of the convolution operation is not optimized towards single matrix multiplications.

\paragraph{Convolution}
The convolution operation, unlike matrix multiplication, does not have a straightforward ``batched'' version.
Instead, we make use of the more general \emph{grouped convolution} operation, which is similar to the original convolution operation but has the restriction that each output channel is calculated from only a certain group of input channels rather than all input channels.

In the context of operation merging, we discovered that holding several groups that are confined from each other is congruent with having isolated input-weight pairs.
Each group corresponds to an input-weight pair.
We merge two convolutions by concatenating the inputs along the channel dimension (same for weights as well), then placing a grouped convolution that consists of a number of groups equal to the number of merged convolutions (i.e., the number of input-weight pairs).
A formal derivation showing that a grouped convolution can produce the exact same results as a set of ordinary convolutions is given in Appendix~\ref{sec:conv-deriv}.

\paragraph{Layer normalization}
Layer normalization~\citep{layernorm} instances can be merged into a single \emph{group normalization}.
Because all input channels are aggregated and normalized at once for layer normalization, simply concatenating the inputs and then using a larger layer normalization instance does not suffice; separate sets of inputs would not be isolated from each other.
Instead, we turn our eyes toward another normalization method, group normalization~\citep{groupnorm}, that breaks up channels into disjoint groups, akin to grouped convolution.
This enables merging layer normalization instances in a manner similar to convolution; we concatenate the inputs and weights along the channel dimension to generate a large input tensor and a large weight tensor, and create a group normalization instance with a number of groups equal to the number of merged layer normalizations.

\paragraph{Operations with input-weight local computations}
The general counterpart operations mentioned in the previous sections -- i.e., batch matrix multiplication, grouped convolution, and group normalization -- can be merged without changing the operation type.
Since these operations allow input-weight local computations by nature, multiple instances of such operations can be merged by concatenating inputs and weights, and increasing the number of local computation groups.
For example, merging $4$ grouped convolutions that use $2$ groups each would result in a large grouped convolution of $4 \times 2 = 8$ groups.
Batch normalization~\citep{batchnorm} can also be merged without special manipulations; the calculations of batch normalization are done in a per-channel manner, so inputs and weights just need to be concatenated along the channel dimension.

\paragraph{Non-trainable operations}
All non-trainable operations can be merged seamlessly, as there are no weights to be merged.
This includes activation functions (e.g., ReLU, Swish, Tanh), max-pooling, mean-pooling, and also other element-wise operations such as plain addition or multiplication.

\subsection{End-to-end DNN Merging} \label{subsec:tech-e2e}
We now extend our discussion to entire DNNs.
More specifically, we show how we merge multiple DNNs that share the same architecture -- i.e., the same sequence of operations -- but incorporate different inputs and different weights.
DNNs can be merged by first merging operations independently, and then reshaping and transposing intermediate tensors between merged operations if necessary.
Whether to add reshaping and transposing operations or not depends on the tensor dimension that operation merging was done.

\begin{figure}[t]
  \includegraphics[width=0.9\textwidth]{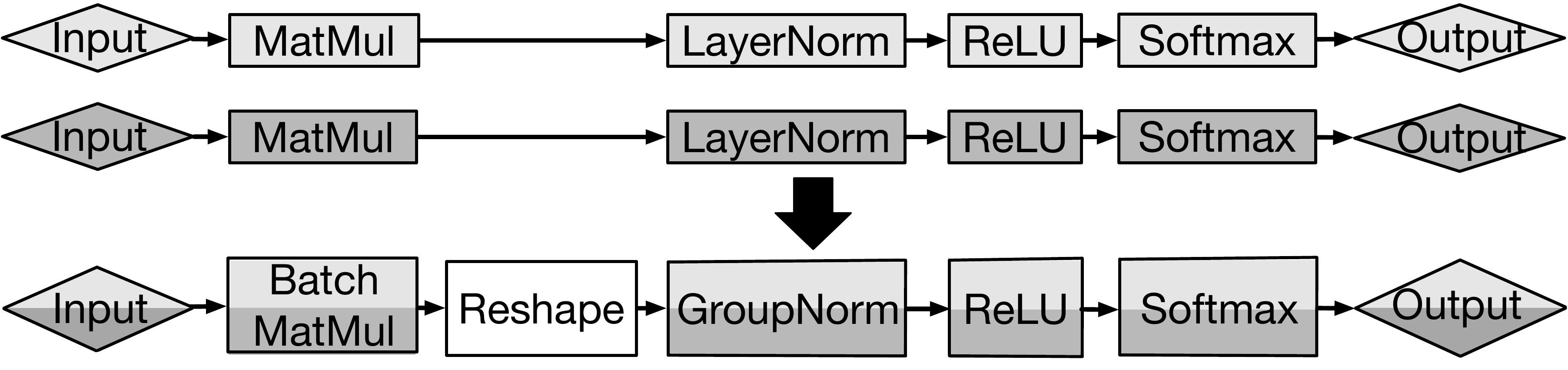}
  \centering
  \caption{An example of merging two FFNNs.
  Individual operations are merged as according to Section~\ref{sec:tech}.
  In case of shape inconsistencies regarding merge dimensions (batch or channel), we insert reshape operations to fix the shapes.}
  \label{fig:mlp-e2e}
\end{figure}

We demonstrate DNN merging with an example of a classical feedforward neural network (FFNN) consisting of a fully connected layer (i.e., matrix multiplication) followed by a layer normalization layer.
Figure~\ref{fig:mlp-e2e} shows two instances of a basic FFNN that share the same network architecture, but contain different weights and serve different inputs (depicted by the difference in shades).
First, the matrix multiplications can be merged into a batch matrix multiplication, given that the inputs are correctly concatenated along the batch dimension.
Next, the layer normalizations can be merged into a group normalization of two groups, but the fact that the previously merged operation produces tensors packed along the batch dimension conflicts with group normalization's merging condition of tensors being concatenated along the channel dimension.
Therefore, we insert a reshaping operation between the previous batch matrix multiplication and the new grouped normalization to ensure that the input tensor of the grouped normalization does indeed have the expected tensor shape.
The rest of the operations in the network are all non-trainable operations, and thus do not require any particular reshaping.

We formally describe the merging process in Algorithm~\ref{alg:dnn}.
Basically, the algorithm is a BFS graph traversal algorithm with a time complexity of $O(|V|+|E|)$.
Visiting each operation, we first check the required merge dimension of the operation (lines 12-16).
Merging matrix multiplications to batch matrix multiplication demands tensors to be concatenated along the $\mathtt{Batch}$ dimension, while grouped convolution, layer normalization, and batch normalization demand concatenation along the $\mathtt{Channel}$ dimension.
Non-trainable operations do not require a specific concatenation scheme, hence we set the dimension as $\mathtt{DontCare}$.
Next, we check whether the merge dimension of the operation is compatible with that of the parent operations (lines 29-31).
If the merge dimensions are not compatible (either $d_{j} = \mathtt{Batch}, d_{i} = \mathtt{Channel}$ or vice versa), then we insert a reshape operation before the merged operation (lines 32-36).

\begin{figure}[h]
\begin{minipage}{\textwidth}
\begin{algorithm}[H]
  \caption{DNN Merging}
  \label{alg:dnn}
  \small
\begin{algorithmic}[1]
    \STATE {\bfseries Input:}  A common subgraph $(V, E)$ of $M$ DNNs and weights $\{w_{ij} \mid w_{ij}$ is the weight parameter for $op_i \in V$ in DNN$_j$ $\}$ 
    \STATE {\bfseries Output:} A graph $(V_{merge}, E_{merge})$ for the merged DNN
    \STATE
    \STATE // start with root ops, with no incoming edges
    \STATE $Q \gets $ empty queue
    \STATE $Q.\mathtt{EnqueueAll}(\{op \in V \mid \not\exists (*, op) \in E\})$
    \STATE
    \WHILE {{\bfseries not} $Q.\mathtt{IsEmpty}()$}
      \STATE $op_{i} \gets Q.\mathtt{Dequeue}()$
      \STATE $\mathtt{MarkVisited}(op_{i})$
      \STATE
      \STATE // $m_i$ indicates the merged op
      \STATE // $d_i$ indicates the concat dimension of tensors
      \STATE // $d_i \in \{ \mathtt{Batch}, \mathtt{Channel}, \mathtt{DontCare} \}$
      \STATE $m_i, d_i \gets \mathtt{Merge}(op_{i}, \{w_{ij}\}_{j=1}^M)$
      \STATE $V_{merge}.\mathtt{Add}(m_i)$
      \STATE
      \STATE // copy op dependencies
      \FOR {$op_j \in \{op_j \in V \mid (op_j, op_i) \in E \}$}
        \STATE $E_{merge}.\mathtt{Add}((m_j, m_i))$
      \ENDFOR
      \STATE
      \IF {$d_i = \mathtt{DontCare}$}
        \STATE // assign the concat dimension of the parent ops
        \STATE // follow the majority if there is a dissensus
        \STATE $d_i \gets \mathtt{MostFrequent}(\{d_j \mid (m_j, m_i) \in E_{merge} \})$ 
      \ENDIF
      \STATE
      \FOR {$m_j \in \{m_j \in V_m \mid \exists (m_j, m_i) \in E_{merge} \}$}
        \STATE // insert reshape op if dimensions are different
        \IF {$d_j \neq d_i$}
          \STATE $E_{merge}.\mathtt{Delete}((m_j, m_i))$
          \STATE $r \gets \mathtt{ReshapeAndTransposeOp}(d_j, d_i)$
          \STATE $E_{merge}.\mathtt{Add}((m_j, r))$
          \STATE $E_{merge}.\mathtt{Add}((r, m_i))$
        \ENDIF
      \ENDFOR
      \STATE
      \STATE $childOps \gets \{op \mid (op_i, op) \in E$ {\bfseries and} $\mathtt{NotVisited}(op)\}$
      \STATE $Q.\mathtt{EnqueueAll}(childOps)$
    \ENDWHILE
    \STATE {\bfseries return} ($V_{merge}, E_{merge}$)
\end{algorithmic}
\end{algorithm}
\end{minipage}
\end{figure}

Although we focused our discussion on DNNs of the exact same architecture, \system is also applicable to DNNs that share common backbones.
In such cases, we merge only the common backbones via Algorithm~\ref{alg:dnn}, and do not merge the other layers.

\section{Implementation}
We have implemented \system as an automated tool on PyTorch 1.3.1~\cite{pytorch}.
\system receives a computation graph (widely employed by modern DL frameworks, including PyTorch) of a DNN model and the number of model instances to merge as input, and outputs a merged version of the computation graph.
Instead of using the common PyTorch model format (\texttt{nn.Module}) as-is, we prepare a PyTorch model in the form of a Torchscript graph, which can be generated from \texttt{nn.Module}s via the Torchscript API.
The Torchscript graph format allows us to implement the graph traversal of Algorithm~\ref{alg:dnn}, whereas \texttt{nn.Module}s do not due to the imperative programming model of PyTorch.
The overall merging mechanism follows the process described in Section~\ref{subsec:tech-e2e}.

The merging process occurs only once per model, offline.
At inference time, the merged model can be run repeatedly without having to go through the merging process again.
In other words, the time overhead for merging models can be amortized across multiple runs.
The largest merging overhead we observed during our experiments was 600 milliseconds for merging 32 ResNeXt-50 instances.
The overhead mostly comes from graph traversal, and does not scale linearly with the number of model instances.

Some Torchscript details require us to treat certain operations with specific measures.
Torchscript considers convolution operations and grouped convolution operations as the same type, \texttt{aten::\_convolution}, and differentiates them by the corresponding integer attribute value for the number of convolution groups.
In other words, a normal convolution op is of \texttt{aten::\_convolution} type with the value 1 for the \texttt{num\_groups} attribute, while a grouped convolution op is of \texttt{aten::\_convolution} type with the number of convolution groups for the \texttt{num\_groups} attribute.
Thus, when converting a convolution op into a grouped convolution op, we don’t actually change the op type, but rather adjust the \texttt{num\_groups} attribute instead.

On the other hand, matrix multiplication operations do not always share the same type as batch matrix multiplications.
PyTorch provides several ways to define a matrix multiplication operation (\texttt{aten::addmm}, \texttt{aten::baddbmm}, etc.).
Therefore, depending on which PyTorch interface is used, it may or may not be possible to convert a matrix multiplication op into a batch matrix multiplication op with a simple tweak in attribute values.
When a type conversion is needed, we not only change the op type but also rewire inputs and outputs, according to the op signature.

\section{Experimental Results} \label{sec:eval}
In this section, we evaluate \system by measuring the inference time of DNNs merged via \system while varying the number of DNNs, the DNN model, the inference batch size, and GPU hardware.
We also check the memory usage of DNN inference and perform other experiments to further understand the characteristics of \system.
\system does not alter the computation results in any way and thus inference accuracy is not affected by our technique.

\subsection{Evaluation Setup}
\textbf{Environment.}
We implemented \system on PyTorch 1.3.1 and used NVIDIA CUDA 10.0 and cuDNN 7.6 to run GPU kernels.
We use an AWS EC2 p3.2xlarge instance, which includes an NVIDIA V100 GPU.
We also use an NVIDIA TITAN Xp on our server machine of two 18-core Intel Xeon E5-2695 @ 2.10 GHz processors with 256GB RAM, for whose experiment results are shown in Appendix~\ref{sec:eval-titanxp}.

\textbf{Models.}
We first experiment on ResNet-50~\citep{resnet} and ResNeXt-50~\citep{resnext}, representative CNNs that are widely used in computer vision.
As ResNet and ResNeXt are mainly used for image classification, we replace the final layer with a fully connected layer of varying output classes to correctly represent multiple classification tasks that have all undergone their own fine-tuning processes.
Excluding the final fully connected layer, all other layers can be merged via \system.
We use synthetic 224x224 RGB images as inputs.

We also perform experiments on BERT~\citep{bert} and XLNet~\citep{xlnet} as representatives of natural language processing models.
Following the paper's guidelines, we run inference tasks by feeding the output of BERT and XLNet to additional fully connected layers.
Each type of task (e.g., question-and-answering, named entity recognition, and sentence/token classification) is associated with its own inputs and number of outputs.
The models themselves are merged via \system.
We use synthetic embeddings of length 128 as inputs.

\textbf{Baselines.}
As stated in Section~\ref{sec:back}, no existing system attempts to merge computations of multiple DNNs when both inputs and weights are completely different.
We implement various baselines on PyTorch that represent a serving system's behavior and compare \system with the baselines:
\begin{itemize}
  \item Sequential: Selects a DNN from the given DNNs in a round-robin fashion and performs inference on each DNN \textit{one by one}.
  \item Concurrent: Assigns \textit{a process per DNN} and lets the processes perform inference on their corresponding DNN without any synchronization across other processes.
  \item Hybrid: Concurrently runs as many models as the GPU memory allows, and then sequentially runs remaining models in the next batches.
\end{itemize}

\begin{figure*}[t]
  \begin{subfigure}[b]{.245\textwidth}
    \includegraphics[width=1.0\linewidth]{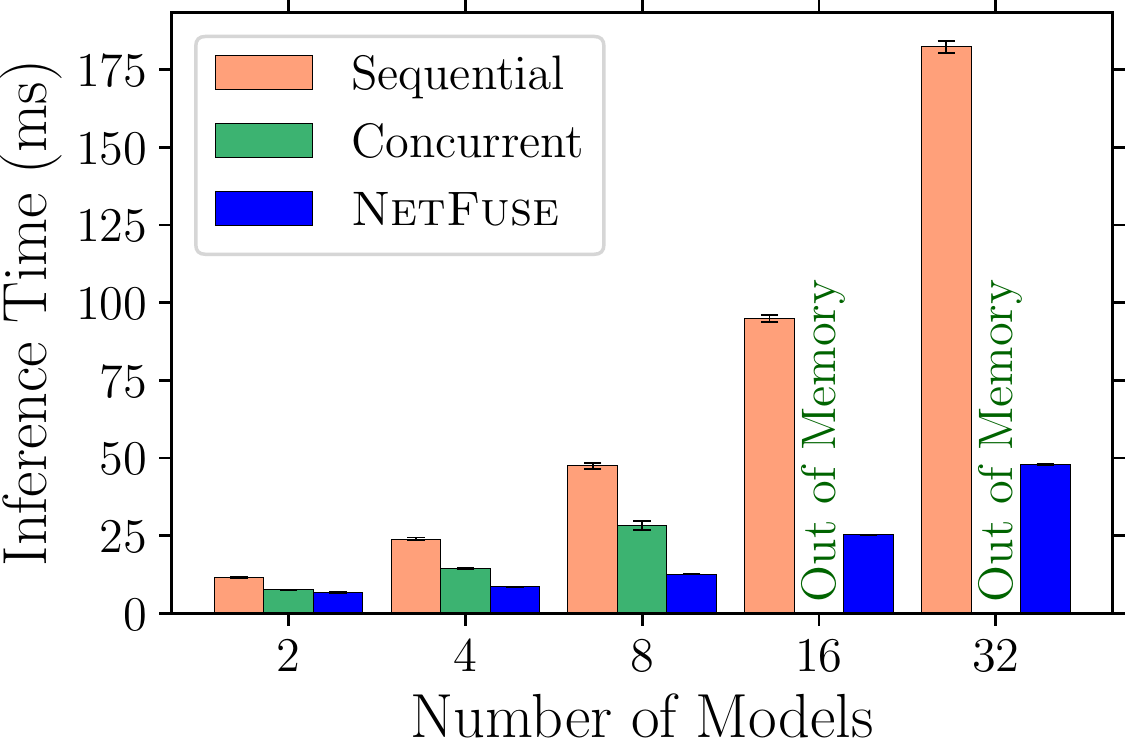}
    \caption{ResNet-50}
    \label{fig:v100_resnet_inftime}
  \end{subfigure}
  \hfill
  \begin{subfigure}[b]{.245\textwidth}
    \includegraphics[width=1.0\linewidth]{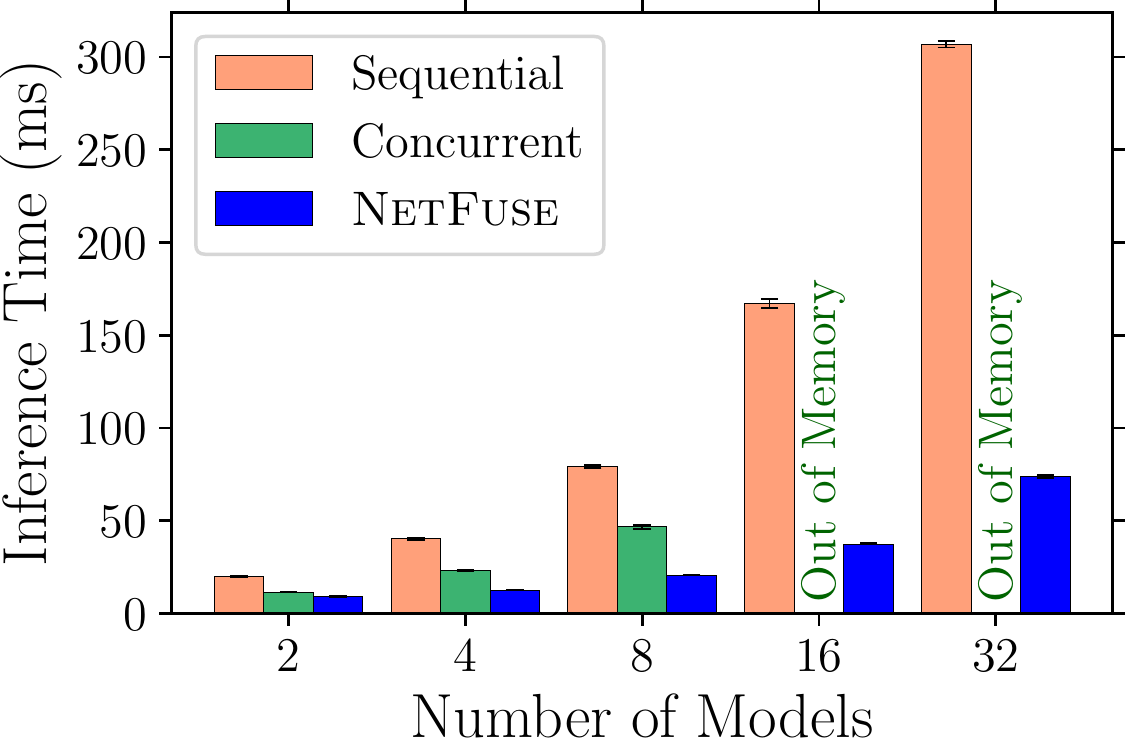}
    \caption{ResNeXt-50}
    \label{fig:v100_resnext_inftime}
  \end{subfigure}
  \hfill
  \begin{subfigure}[b]{.245\textwidth}
    \includegraphics[width=1.0\linewidth]{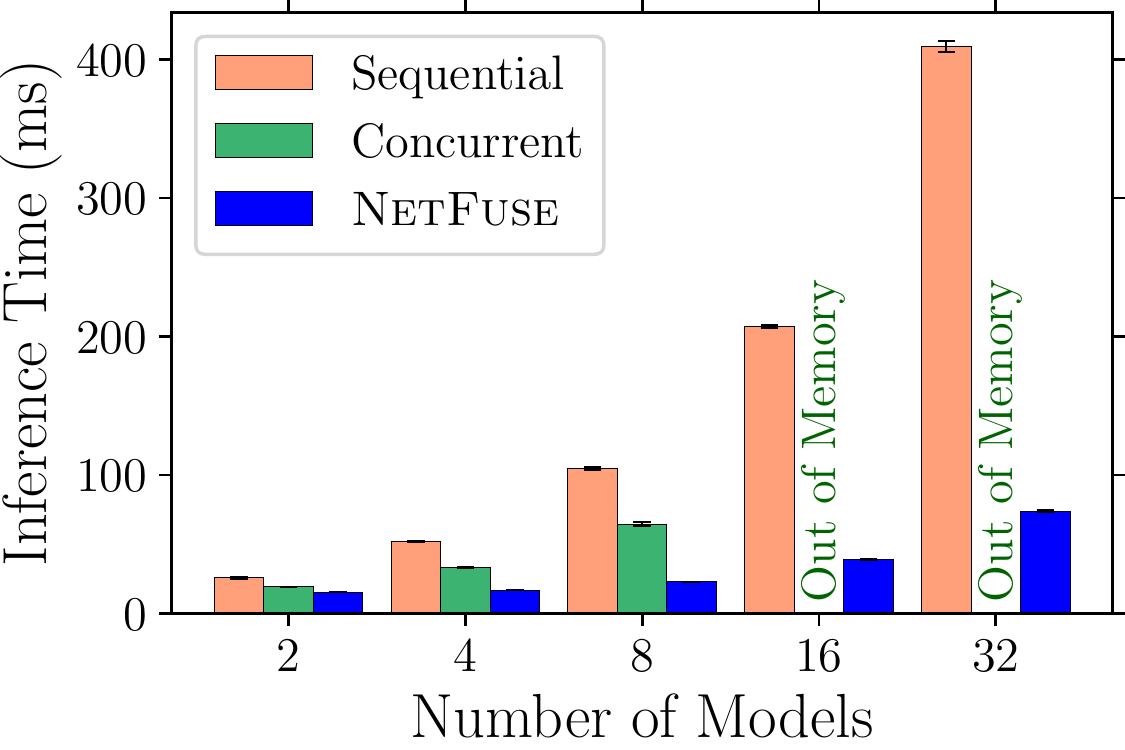}
    \caption{BERT}
    \label{fig:v100_bert_inftime}
  \end{subfigure}
  \hfill
  \begin{subfigure}[b]{.245\textwidth}
    \includegraphics[width=1.0\linewidth]{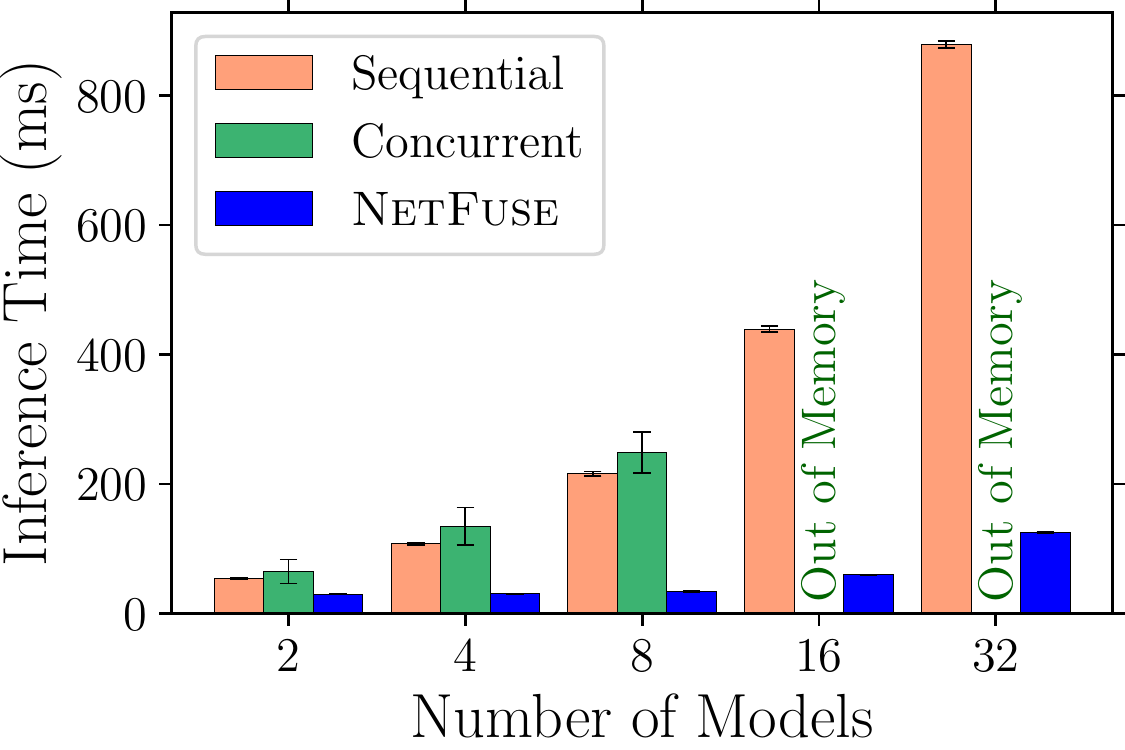}
    \caption{XLNet}
    \label{fig:v100_xlnet_inftime}
  \end{subfigure}
  \centering
  \caption{Mean inference time of \system and the sequential and concurrent baselines for a varying number of models on V100. The batch size is set to $1$. The error bars indicate the standard deviation.}
  \label{fig:v100_inftime}
\end{figure*}

\begin{figure*}[t]
    \centering
    \includegraphics[width=\textwidth]{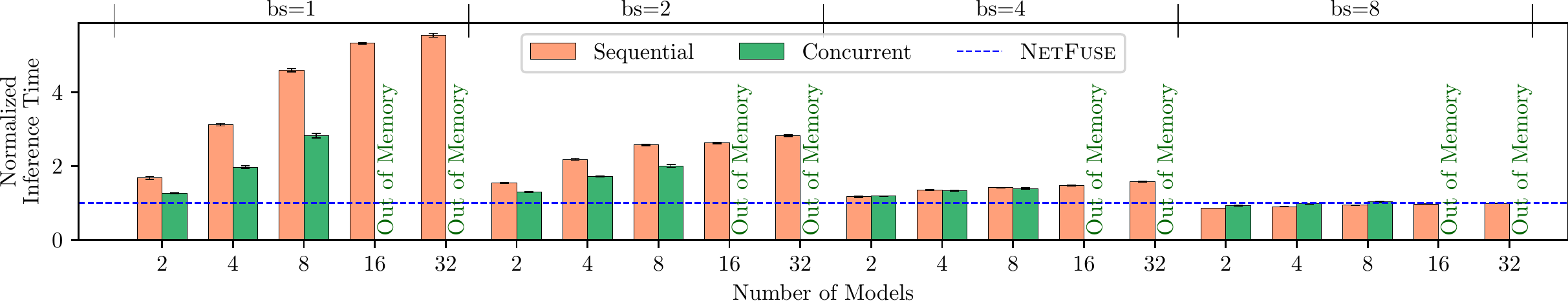}
  \centering
  \caption{Inference time of \system and the sequential and concurrent baselines on BERT, normalized by the inference time of \system, for a varying number of batch sizes (\textbf{bs}) on V100. The error bars indicate the standard deviation.}
  \label{fig:v100_bert_normalized_inftime}
\end{figure*}

\subsection{Inference Time} \label{subsec:eval-inftime}
Figure~\ref{fig:v100_inftime} presents the inference time of performing inference on various models for \system and the two baselines, on a V100 GPU.
Each bar represents the mean inference time of 1,000 runs for the corresponding configuration.

Experiment results on ResNet-50 are depicted on Figure~\ref{fig:v100_resnet_inftime}.
As we increase the number of models, the inference time of the sequential baseline grows linearly because it must sequentially process each inference without any overlapping any computations across models.
The concurrent baseline performs better than the sequential baseline as the GPU is being fed with more requests, but fails to reach the speed of \system because the computations of different models are still being launched as independent kernels.
In fact, the concurrent baseline runs out of GPU memory for large numbers of models (explained in Section~\ref{subsec:eval-memory}).
On the other hand, \system is able to merge computations of different models and achieve lower inference time then other baselines.
A similar trend is repeated for ResNeXt-50 in Figure~\ref{fig:v100_resnext_inftime} and BERT in Figure~\ref{fig:v100_bert_inftime}.
Interestingly, the concurrent baseline is the slowest for XLNet, as shown in Figure~\ref{fig:v100_xlnet_inftime}.
We conjecture that the extra computations used in XLNet's base architecture, Transformer-XL, compared to BERT's base architecture, Transformer, renders concurrent executions more ineffective.
The inference time speedup is up to 2.6$\times$, 3.4$\times$, 2.7$\times$, 3.6$\times$ for ResNet-50, ResNeXt-50, BERT, and XLNet, respectively.

In order to examine how batch size affects \system's effectiveness, we repeated the previous experiment for BERT on greater batch sizes and draw the results in Figure~\ref{fig:v100_bert_normalized_inftime}.
The inference times of the baselines are shown as relative numbers against \system (the blue dotted horizontal line 1x).
Although \system is faster than the other baselines for most configurations, the gap between \system and the baselines gradually decreases as the batch size increases.
There even exists a configuration (batch size 8) where \system performs more poorly than the baselines.
This is because the GPU is already well saturated with a large batch size, and thus further merging computations does not affect the GPU's utilization enough to improve speed.
Nevertheless, \system performs significantly better than other baselines for small batch sizes and does not experience GPU memory issues like the concurrent baseline.

\begin{figure*}[h]
  \begin{subfigure}[b]{.245\textwidth}
    \includegraphics[width=1.0\linewidth]{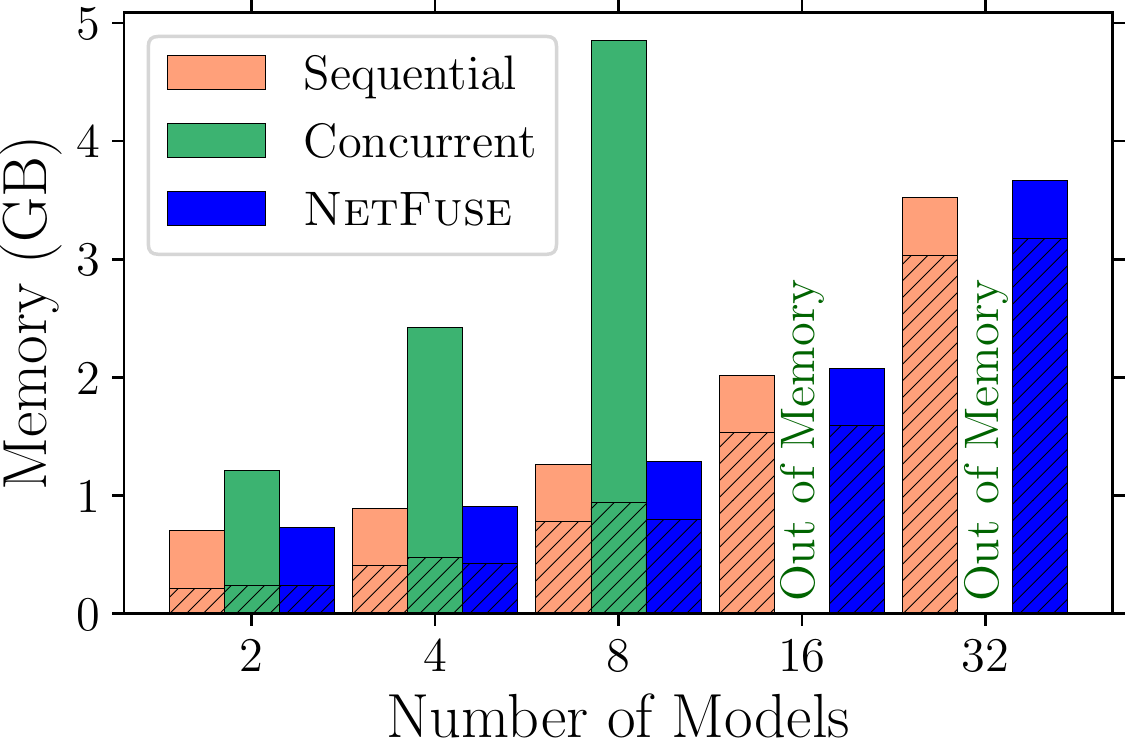}
    \caption{ResNet-50}
    \label{fig:v100_resnet_memory}
  \end{subfigure}
  \begin{subfigure}[b]{.245\textwidth}
    \includegraphics[width=1.0\linewidth]{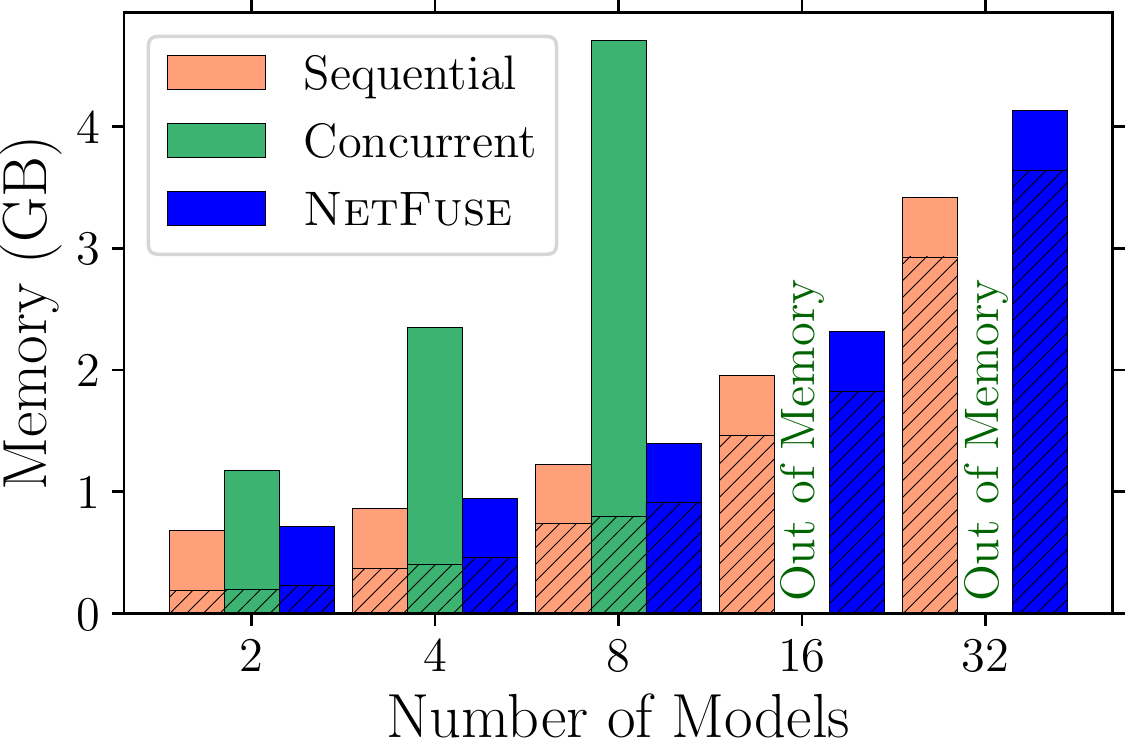}
    \caption{ResNeXt-50}
    \label{fig:v100_resnext_memory}
  \end{subfigure}
  \begin{subfigure}[b]{.245\textwidth}
    \includegraphics[width=1.0\linewidth]{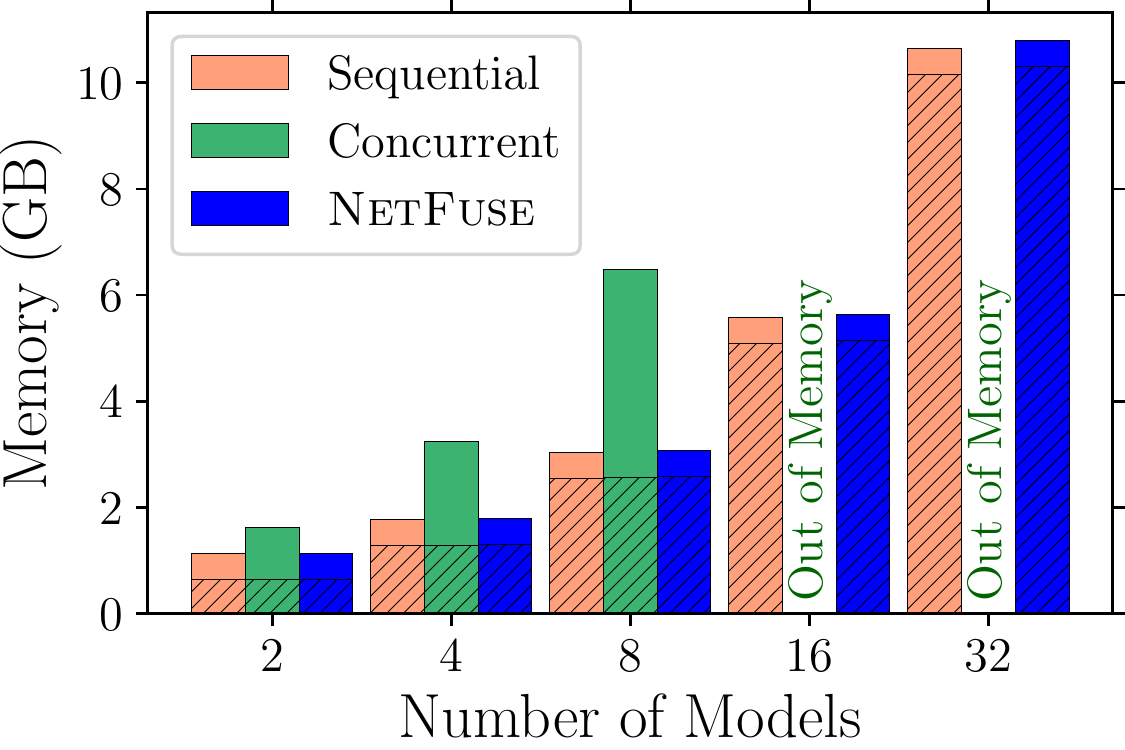}
    \caption{BERT}
    \label{fig:v100_bert_memory}
  \end{subfigure}
  \begin{subfigure}[b]{.245\textwidth}
    \includegraphics[width=1.0\linewidth]{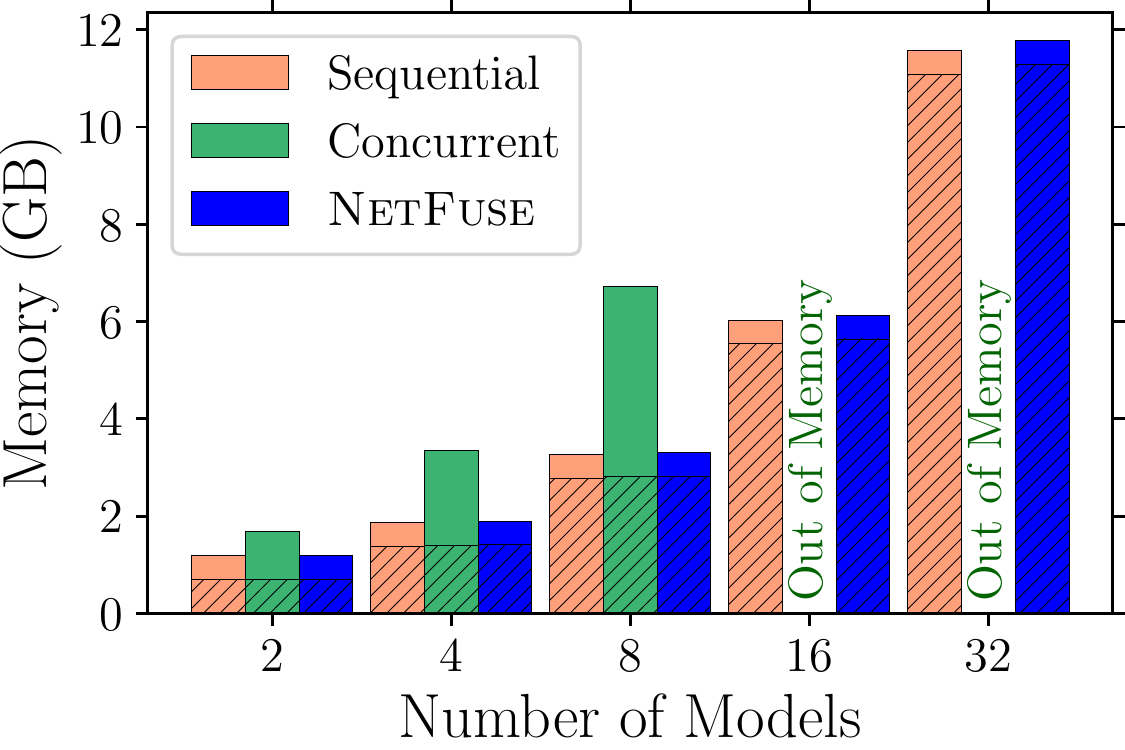}
    \caption{XLNet}
    \label{fig:v100_xlnet_memory}
  \end{subfigure}
  \centering
  \caption{Peak GPU memory usage of \system and the sequential and concurrent baselines for a varying number of models on V100. The batch size is set to $1$. For each vertical bar,
  the hatched portion denotes the amount of memory used as inference workspace to house weights and intermediate activation values, while the solid portion corresponds to the base memory reserved by the framework, PyTorch (500 MBs per process).
  V100 has a total amount of 16 GB memory.}
  \label{fig:v100_memory}
\end{figure*}

\subsection{Memory Footprint} \label{subsec:eval-memory}
We further investigate the GPU memory issue of the concurrent baseline by measuring the peak GPU memory usage of \system and the baselines during inference.
In Figure~\ref{fig:v100_memory}, we show the maximum amount of memory used by \system and the baselines for different configurations.
The hatched portion of each bar indicates the amount of memory used as inference workspace (weights and activations), while the remaining solid portion of each bar indicates the base memory held by the framework, PyTorch.
The main reason for the concurrent baseline running out of memory is the base memory, as PyTorch takes $500$ MBs per process when using the GPU.
Spawning $16$ processes to serve $16$ models results in PyTorch taking approximately $8$ GBs, which is already half of the V100 GPU's total memory, $16$ GBs.
Additionally, the memory used by the sequential baseline is the smallest for all cases because the sequential baseline performs only one model's worth of inference at a time, unlike the concurrent baseline and \system.

\begin{figure*}[t]
    \centering
    \begin{subfigure}[b]{.287\textwidth}
    \centering
    \includegraphics[width=\linewidth]{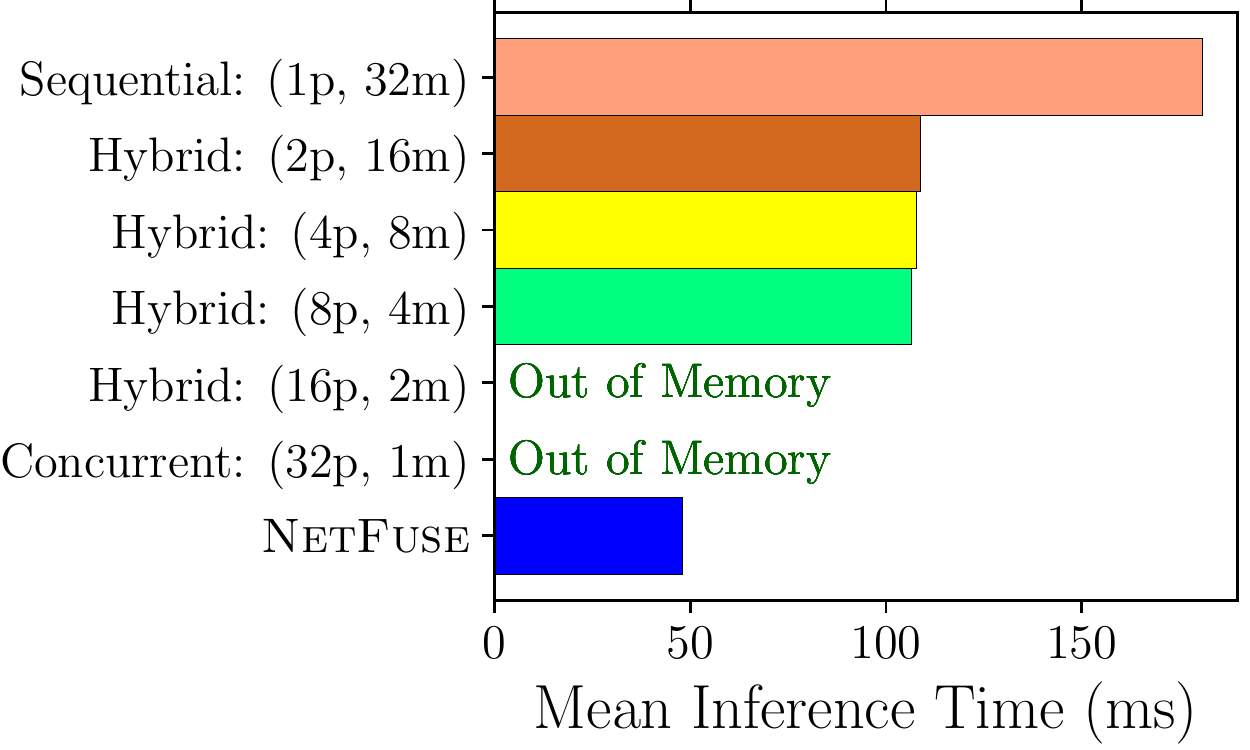}
    \caption{ResNet-50}
    \label{fig:hybrid_resnet}
    \end{subfigure}
    \hfill
    \begin{subfigure}[b]{.23\textwidth}
    \centering
    \includegraphics[width=\linewidth]{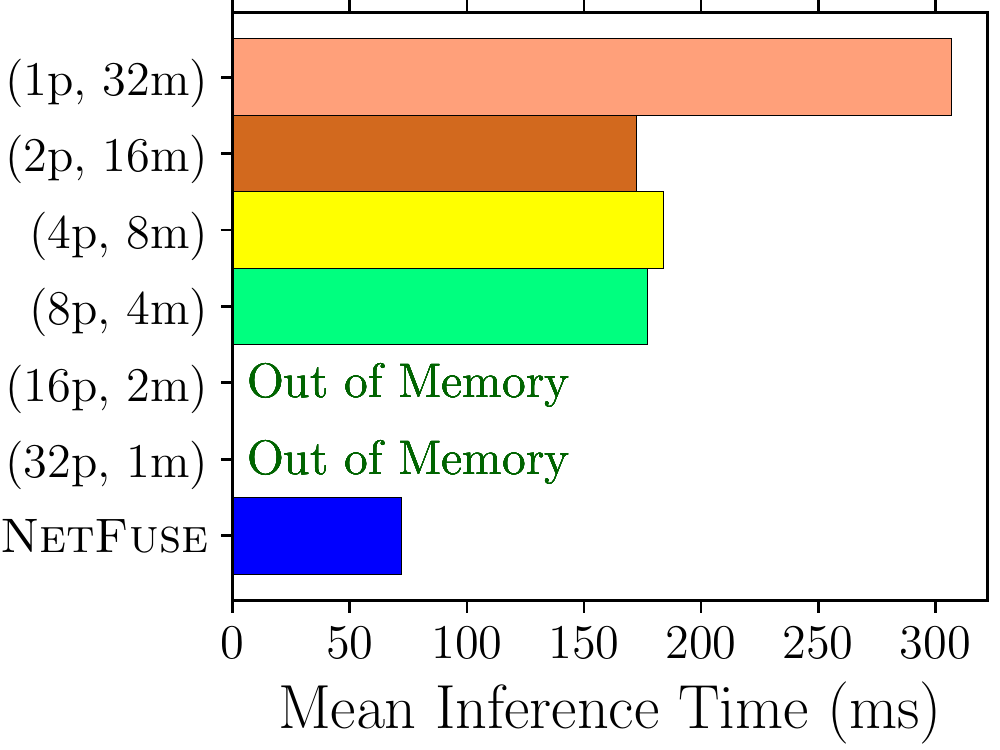}
    \caption{ResNeXt-50}
    \label{fig:hybrid_resnext}
    \end{subfigure}
    \hfill
    \begin{subfigure}[b]{.23\textwidth}
    \centering
    \includegraphics[width=\linewidth]{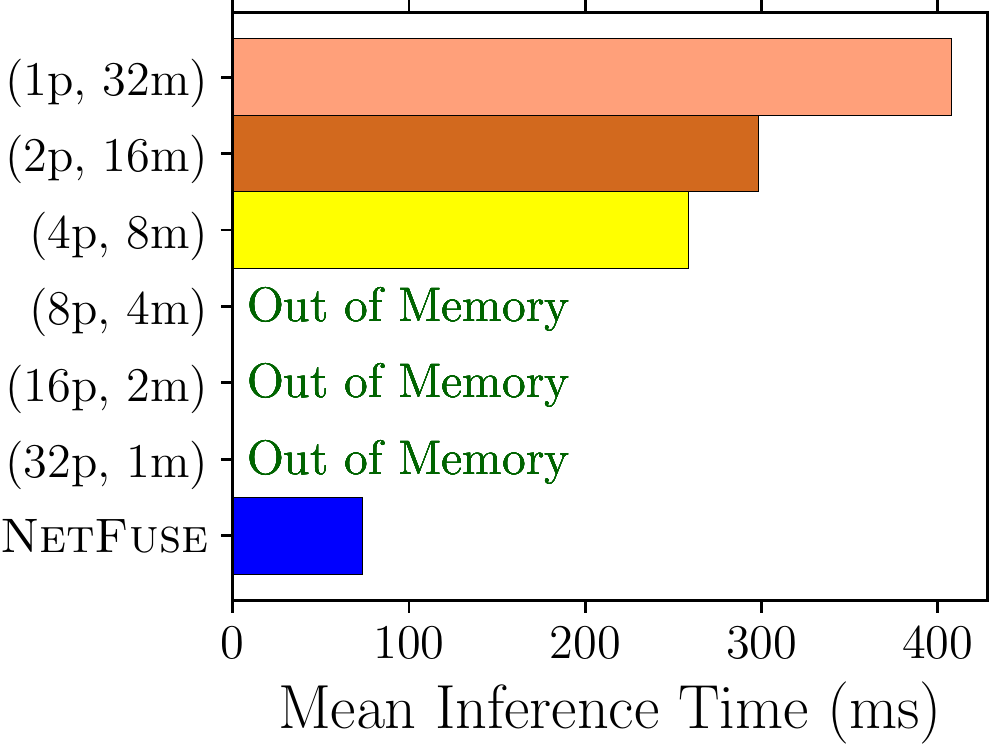}
    \caption{BERT}
    \label{fig:hybrid_bert}
    \end{subfigure}
    \hfill
    \begin{subfigure}[b]{.23\textwidth}
    \centering
    \includegraphics[width=\linewidth]{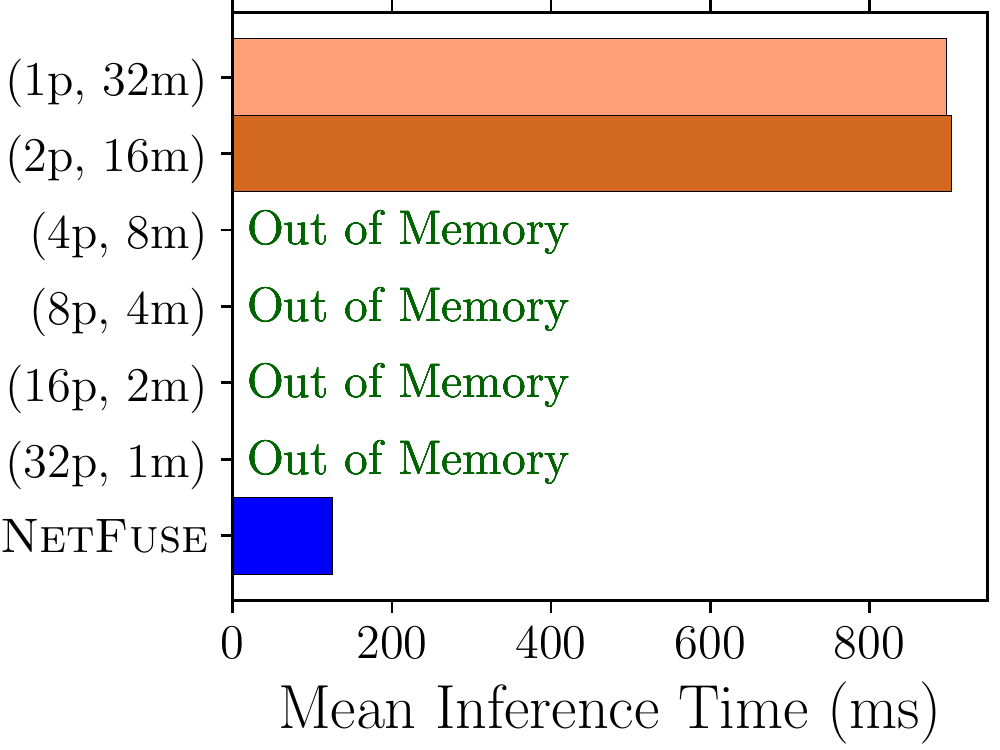}
    \caption{XLNet}
    \label{fig:hybrid_xlnet}
    \end{subfigure}
  \caption{Mean inference time of \system and the sequential, concurrent, and hybrid baselines on V100. The batch size is set to $1$. $(Ap, Bm)$ denotes a configuration of $A$ concurrent processes, each running $B$ models sequentially.}
  \centering
  \label{fig:hybrid}
\end{figure*}

\textbf{Sequential-concurrent hybrid strategy.}
The concurrent baseline generally tends to be faster than the sequential baseline, but suffers from memory issues for large numbers of models.
Naturally, one can think of a hybrid approach that combines the strengths of the concurrent and sequential baselines - spawn concurrent processes (per model) as much as the GPU memory allows, and make each process run a number of models sequentially.
For instance, instead of creating $32$ processes to serve $32$ models like the concurrent baseline, we can generate $4$ processes that run $8$ models each.

Figure~\ref{fig:hybrid} shows the inference time of this hybrid approach for running $32$ models, along with the other baselines and \system.
While the concurrent baseline runs out of memory, the hybrid approach is able to avoid this issue by spawning less processes.
As an example, we can see in Figure~\ref{fig:hybrid_resnet} that the hybrid configurations of spawning $2$, $4$, and $8$ processes do not run out of memory.
At the same time, they exhibit shorter inference times than the purely seqential baseline by running multiple models concurrently.
Nonetheless, \system still outperforms the hybrid baseline by up to 2.5$\times$ for ResNeXt-50 and 7.2$\times$ for XLNet.
Note that the hybrid approach may still be susceptible to memory issues, depending on the model;
as can be seen in Figure~\ref{fig:hybrid_xlnet}, even a relatively small number of processes leads to running out of memory for XLNet.

\section{Discussion}

\textbf{Applicability of \system on training models.}
\system can be used to train several models as one large model.
As the group operations listed in Table~\ref{tbl:op-list} (grouped convolution, batch matrix multiplication, and group normalization) all have proper backpropagation operations, a merged model can be trained on deep learning frameworks just as ordinary models.
Indeed, we have confirmed that \system brings similar performance benefits to both training and inference.

However, \system may be less effective on training than inference, for a few practical reasons.
First, DNN training is typically done in larger batch sizes than inference (10s to 1000s).
As presented in Figure~\ref{fig:v100_bert_normalized_inftime},
\system's performance degrades when the batch size increases, so \system should be applied selectively to training models of small batch sizes or training small models.
Second, individual models may have different training lengths as well as different hyperparameters and optimizers.
In order to accommodate these factors when training merged models, additional measures are required such as excluding models that have finished training and merging the optimizers of each individual model.

\textbf{Applicability of \system on models of different architectures.}
\system, as well as all previous frameworks~\citep{clipper, pretzel, mcdnn, mainstream, nexus, hivemind} noted in our paper, are not applicable to models with completely different architectures.
For instance, we do not consider merging a CNN with an LSTM;
the two models are structurally too different.
That said, \system is applicable to models that are not quite completely identical, but do have common backbones.
The BERT model is a perfect example - the attention layers are unmodified (only the weights are fine-tuned), but the following fully connected layers are customized depending on the NLP task.
In such cases, we merge the backbones, but leave the customized layers (fully connected layers specific for downstream tasks) as-is.
In fact, this is how we merged the models in our experiments.

\section{Conclusion}
In this paper, we introduce \system, a merging technique that can be applied to merging DNNs that share the same architecture but house different parameters and different inputs.
By finding general counterparts for DNN operations that allow input-weight local computations, \system is able to merge multiple operations into a single large operation while preserving the same outputs.
We also show how merging works for whole DNNs and propose an algorithm for DNN merging.
Our experiments show that \system indeed performs faster than baselines with a small cost of additional GPU memory.

\renewcommand\thesection{\Alph{section}}
\renewcommand\thesubsection{\thesection.\arabic{subsection}}

\setcounter{section}{0}
\renewcommand*{\theHsection}{appendix.\Alph{section}}
\renewcommand*{\theHsubsection}{\theHsection.\arabic{subsection}}

\section*{Appendix}
\section{Convolution Derivation} \label{sec:conv-deriv}

We now formally show that a grouped convolution can produce the exact same results as a set of ordinary convolutions, when merged correctly.
We present our analysis based on 2D convolutions, but we note that this can be readily generalized to both 1D and 3D convolutions.
We first go over the definitions of the convolution and grouped convolution operations, and then derive that a grouped convolution of $M$ groups is mathematically equivalent to $M$ convolutions.

Consider a convolution operation which takes a tensor $\boldsymbol{x}$ of shape $(C_{in}, H_{in}, W_{in})$ as input, where $C_{in}$ denotes the number of channels (filter maps) and $H_{in}$, $W_{in}$ denote the height and width, respectively.
Below, we use the notation $\boldsymbol{x}[c]$ to indicate $\boldsymbol{x}$'s $c$-th subtensor of shape $(H_{in}, W_{in})$.
We also define the weight tensor, $\boldsymbol{w}$ of shape $(C_{out}, C_{in}, K, K)$, and the output tensor, $\boldsymbol{y}$ of shape $(C_{out}, H_{out}, W_{out})$, in a similar manner.
$C_{out}$ indicates the number of output channels, $H_{out}$ and $W_{out}$ denote the height and width of the output tensor, and $K$ dictates the kernel size of the weight tensor.
Then, a specific subtensor of $\boldsymbol{y} \coloneqq \mathtt{Conv}(\boldsymbol{x}, \boldsymbol{w})$ can be calculated as follows:
\begin{equation} \label{eq:conv}
\begin{gathered}
  \boldsymbol{y}[c] = \sum_{c_{in}=0}^{C_{in}-1} \boldsymbol{w}[c, c_{in}] \star \boldsymbol{x}[c_{in}]
\end{gathered}
\end{equation}

$\star$ denotes the valid cross-correlation operator for 2D tensors.

Next, a grouped convolution ($G$ convolution groups) of an input tensor $\boldsymbol{x}$ of shape $(C_{in}G, H_{in}, W_{in})$ and a weight tensor $\boldsymbol{w}$ of shape $(C_{out}G, C_{in}, K, K)$ that produces an output tensor $\boldsymbol{y} \coloneqq \mathtt{GroupConv}(\boldsymbol{x}, \boldsymbol{w}, G)$ of shape $(C_{out}G, H_{out}, W_{out})$ can be expressed as:
\begin{equation} \label{eq:groupconv}
\begin{gathered}
  \boldsymbol{y}[c] = \sum_{c_{in}=0}^{C_{in}-1} \boldsymbol{w}[c, c_{in}] \star \boldsymbol{x}[g+c_{in}] \\
  \text{where} \quad g = C_{in}\floor*{\frac{c}{C_{out}}}
\end{gathered}
\end{equation}

The term $\floor*{\frac{c}{C_{out}}}$ in $g$ indicates which convolution group $c$ belongs to.
For example, the channels $c = 0, 1, ..., C_{out}-1$ compose the first convolution group, and thus $\floor*{\frac{c}{C_{out}}} = 0$.
Note that when $G = 1$, this becomes an ordinary convolution, i.e., $\mathtt{GroupConv}(\boldsymbol{x}, \boldsymbol{w}, 1) = \mathtt{Conv}(\boldsymbol{x}, \boldsymbol{w})$.

Finally, we show that it is possible to perform $M$ convolutions with a single grouped convolution operation.
Given $M$ input tensors $\{\boldsymbol{x}_m\}_{m=0}^{M-1}$ ($\boldsymbol{x}_m$ is of shape $(C_{in}, H_{in}, W_{in})$) and $M$ weight tensors $\{\boldsymbol{w}_m\}_{m=0}^{M-1}$ ($\boldsymbol{w}_m$ is of shape $(C_{out}, C_{in}, K, K)$),
we concatenate the input tensors, along the channel dimension, into a large input tensor $\boldsymbol{x}$ of shape $(C_{in}M, H_{in}, W_{in})$.
This way, a specific subtensor of $\boldsymbol{x}$ corresponds to a specific subtensor of $\boldsymbol{x}_m$, as in $\boldsymbol{x}[C_{in}m + c_{in}] = \boldsymbol{x}_m[c_{in}]$.
We repeat this process for the weight tensors as well to create a large weight tensor $\boldsymbol{w}$.

With $\boldsymbol{x}$ and $\boldsymbol{w}$ in hand,
we define $\boldsymbol{y}$ as the output of performing grouped convolution on $\boldsymbol{x}$ and $\boldsymbol{w}$ with $M$ groups.
Considering $\boldsymbol{y}$ has a shape of $(C_{out}M, H_{out}, W_{out})$, we denote the first $C_{out}$ subtensors of $\boldsymbol{y}$ as $\boldsymbol{y}_0$, the second $C_{out}$ subtensors of $\boldsymbol{y}$ as $\boldsymbol{y}_1$, and so on:
\begin{equation} \label{eq:y-subtensor}
\begin{gathered}
  \boldsymbol{y}_m \coloneqq \boldsymbol{y}[C_{out}m:C_{out}(m+1)] \\
  \Longleftrightarrow \boldsymbol{y}_m[c_{out}] = \boldsymbol{y}[C_{out}m+c_{out}] \\
  (m = 0, 1, ..., M-1)
\end{gathered}
\end{equation}

Using Eq.~\ref{eq:groupconv} and Eq.~\ref{eq:y-subtensor}, we can derive the following for a specific subtensor of $\boldsymbol{y}_m$:
\begin{equation} \label{eq:init-formula}
\begin{split}
  \boldsymbol{y}_m[c_{out}]
  &= \boldsymbol{y}[C_{out}m + c_{out}] \\
  &= \sum_{c_{in}} \boldsymbol{w}[C_{out}m + c_{out}, c_{in}] \star \boldsymbol{x}[g+c_{in}] \\
  &= \sum_{c_{in}} \boldsymbol{w}_m[c_{out}, c_{in}] \star \boldsymbol{x}[g+c_{in}]
\end{split}
\end{equation}

Note that all channels $c_{out}$ of $\boldsymbol{y}_m$ correspond to the $m$-th convolution group.
This is confirmable by recalculating $g$ in Eq.~\ref{eq:groupconv} with the fact that $c$ has been replaced with $C_{out}m + c_{out}$ from Eq.~\ref{eq:init-formula} in mind:
\begin{equation} \label{eq:g}
\begin{split}
  g &= C_{in} \floor*{\frac{(C_{out}m + c_{out}) }{C_{out}}} \\
  &= C_{in}\floor*{m + \frac{c_{out}}{C_{out}}} \\
  &= C_{in}m \quad (\because 0 \leq c_{out} < C_{out})
\end{split}
\end{equation}

At last, substituting $C_{in}m$ for $g$ in Eq.~\ref{eq:init-formula} gives us:
\vspace{-15pt}
\begin{proof}[\unskip\nopunct]
\vspace{-10pt}
\begin{align*} \label{eq:final-formula}
  \boldsymbol{y}_m[c_{out}]
  &= \sum_{c_{in}} \boldsymbol{w}_m[c_{out}, c_{in}] \star \boldsymbol{x}[g+c_{in}] \\
  &= \sum_{c_{in}} \boldsymbol{w}_m[c_{out}, c_{in}] \star \boldsymbol{x}[C_{in}m+c_{in}] \numberthis \\
  &= \sum_{c_{in}} \boldsymbol{w}_m[c_{out}, c_{in}] \star \boldsymbol{x}_m[c_{in}]
\end{align*}
$\qquad \Longrightarrow \boldsymbol{y}_m = \mathtt{Conv}(\boldsymbol{x}_m, \boldsymbol{w}_m) \qquad \qquad \qquad$ \qedhere
\end{proof}

Thus, we are essentially performing all $M$ convolutions with one single grouped convolution to evaluate the exact same results, with no redundant nor missing computations.
\section{Experimental Results for NVIDIA TITAN Xp} \label{sec:eval-titanxp}
In this section, we present the results of experiments executed on an NVIDIA TITAN Xp GPU.
Similar to Sections~\ref{subsec:eval-inftime} and~\ref{subsec:eval-memory}, we examined both the inference time and memory footprint of \system and the baselines.

\subsection{Inference Time}
\begin{figure*}[h]
  \begin{subfigure}[b]{.245\textwidth}
    \includegraphics[width=1.0\linewidth]{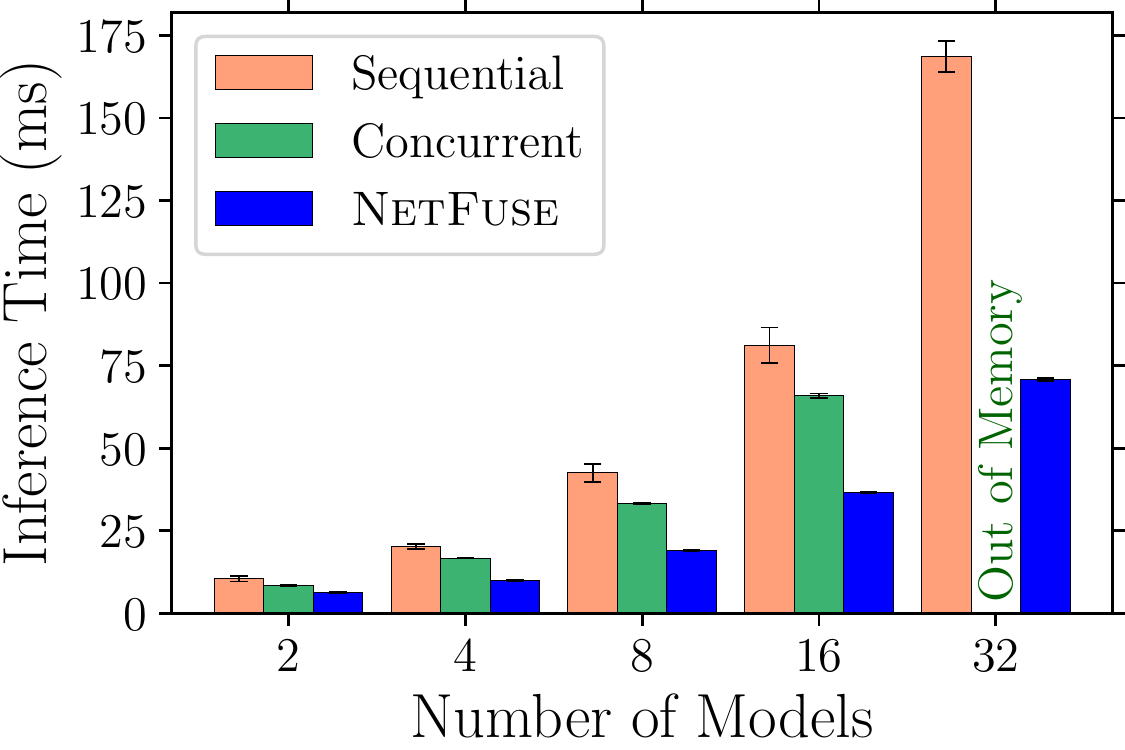}
    \caption{ResNet-50}
    \label{fig:xp_resnet_inftime}
  \end{subfigure}
  \hfill
  \begin{subfigure}[b]{.245\textwidth}
    \includegraphics[width=1.0\linewidth]{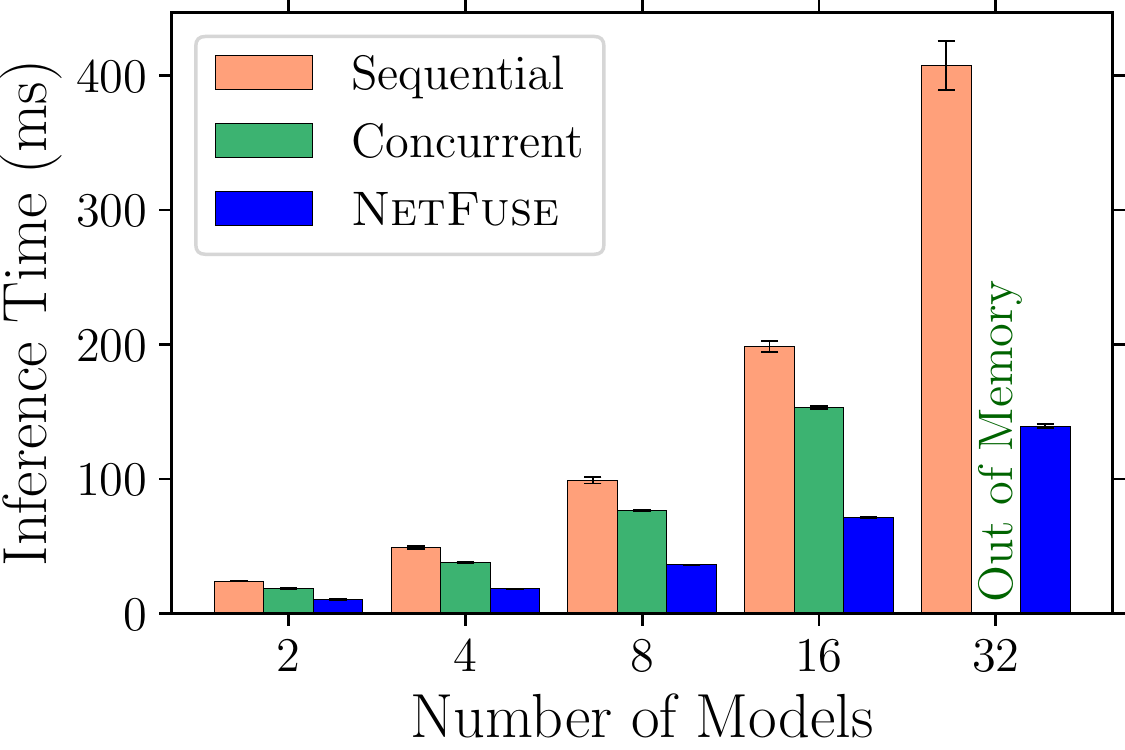}
    \caption{ResNeXt-50}
    \label{fig:xp_resnext_inftime}
  \end{subfigure}
  \hfill
  \begin{subfigure}[b]{.245\textwidth}
    \includegraphics[width=1.0\linewidth]{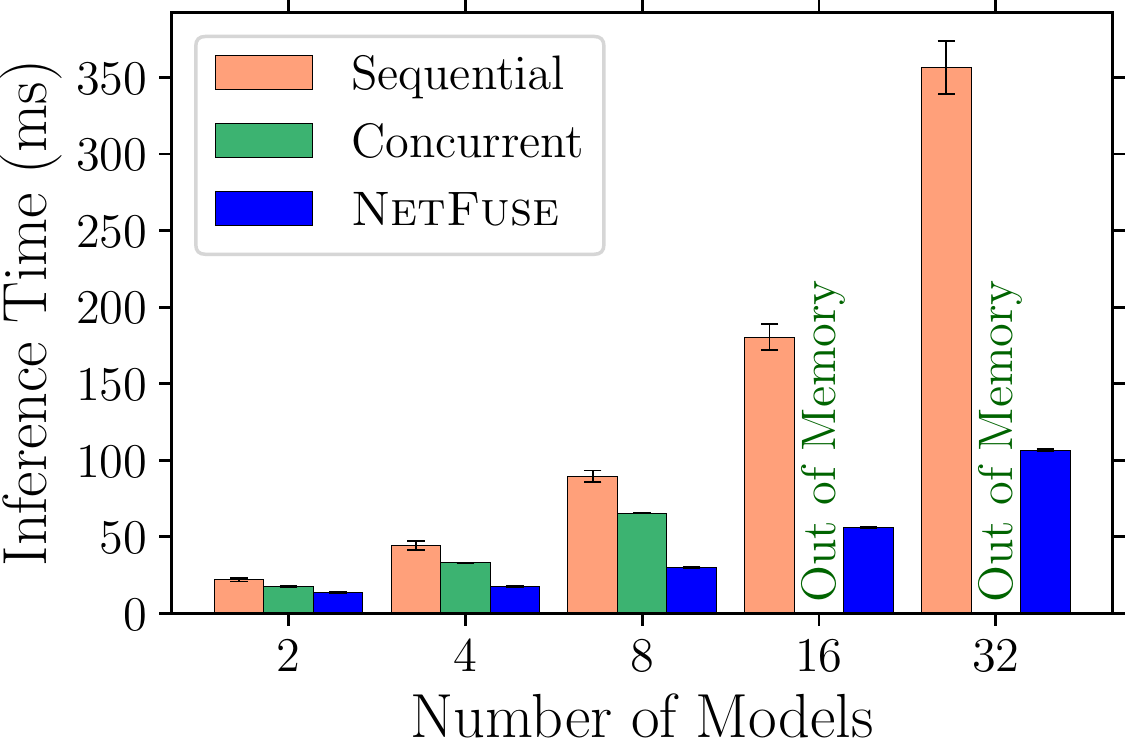}
    \caption{BERT}
    \label{fig:xp_bert_inftime}
  \end{subfigure}
  \hfill
  \begin{subfigure}[b]{.245\textwidth}
    \includegraphics[width=1.0\linewidth]{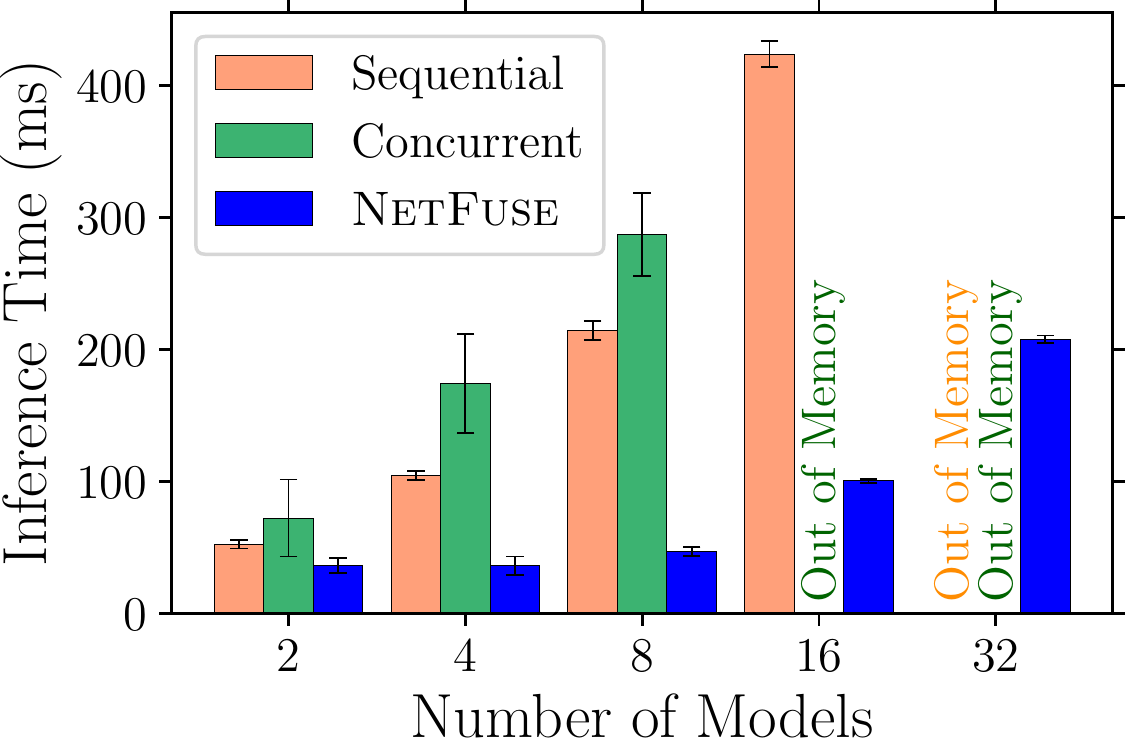}
    \caption{XLNet}
    \label{fig:xp_xlnet_inftime}
  \end{subfigure}
  \centering
  \caption{Mean inference time of \system and the sequential and concurrent baselines for a varying number of models on TITAN Xp. The batch size is set to $1$. The error bars indicate the standard deviation.}
  \label{fig:titanxp_inftime}
\end{figure*}

Figure~\ref{fig:titanxp_inftime} shows the inference time when performing inference for the four models described in Section~\ref{subsec:eval-inftime}.
The height of each bar indicates the mean inference time of 1,000 runs for the corresponding configuration.
The overall trend of \system outperforming the baselines, which was seen in V100 experiments, is also present here as well.
The relative performance gains are lower than the gains on V100, which is due to the fact that the V100 GPU has significantly more cores than the TITAN Xp GPU, and thus can more effectively parallelize the processing of merged models.

\subsection{Memory Footprint}

\begin{figure*}[h]
  \begin{subfigure}[b]{.245\textwidth}
    \includegraphics[width=1.0\linewidth]{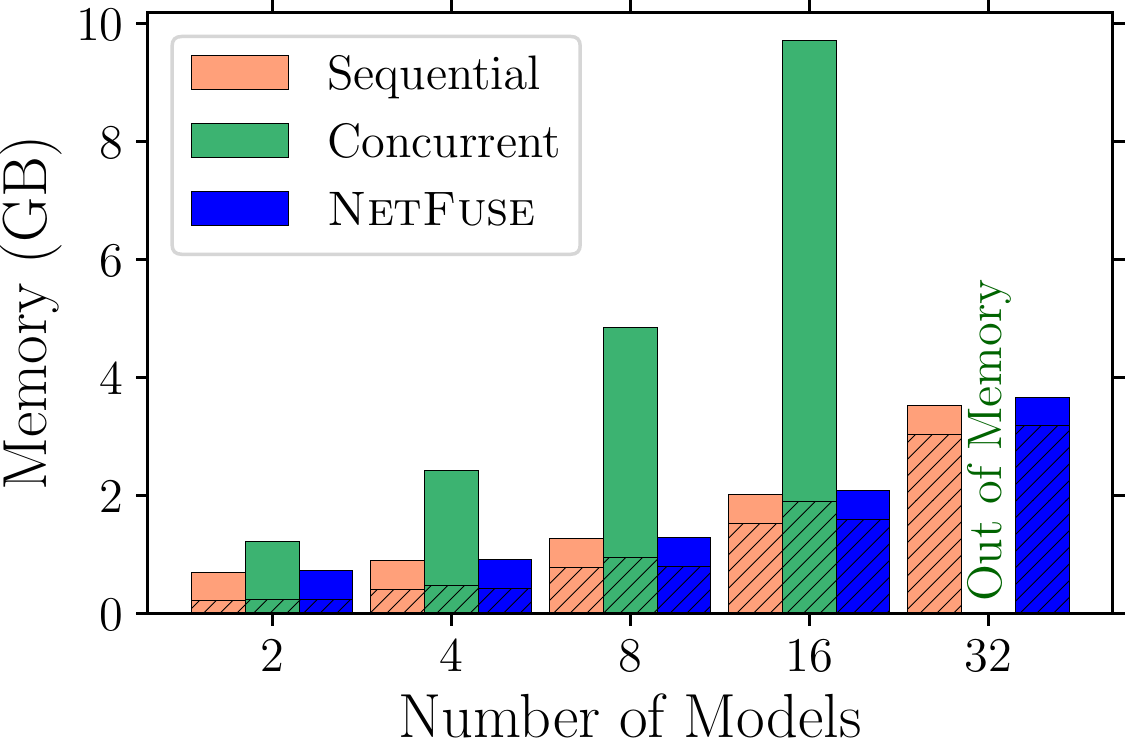}
    \caption{ResNet-50}
    \label{fig:xp_resnet_memory}
  \end{subfigure}
  \hfill
  \begin{subfigure}[b]{.245\textwidth}
    \includegraphics[width=1.0\linewidth]{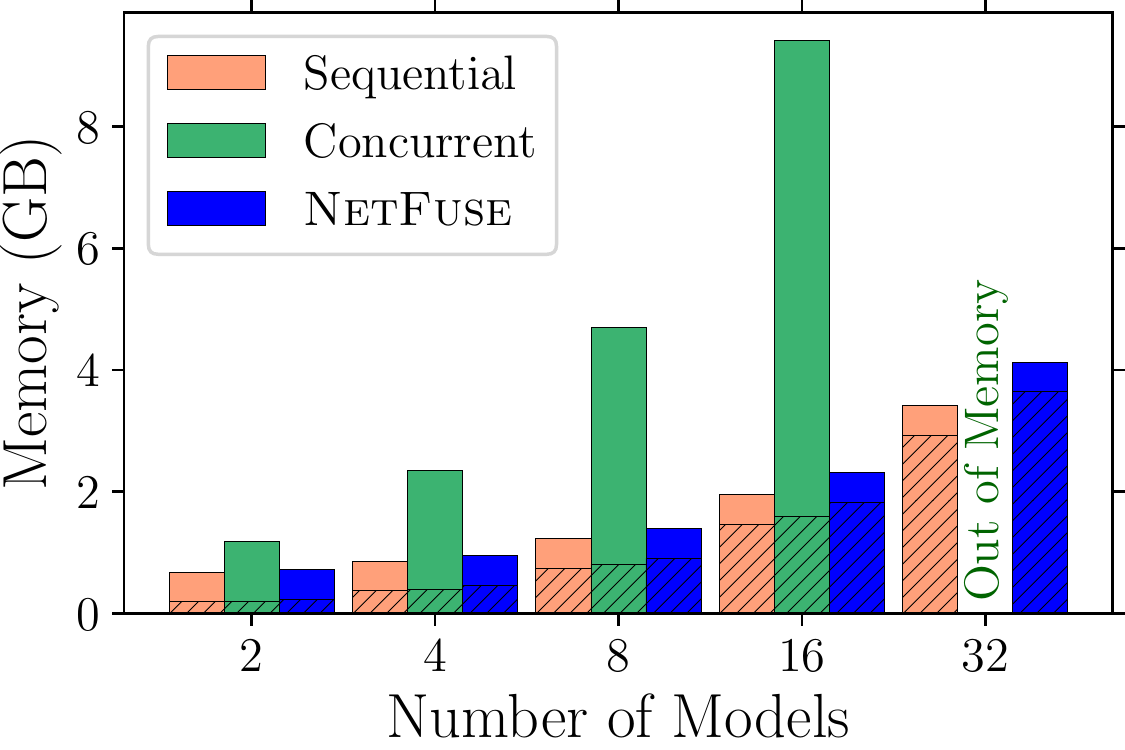}
    \caption{ResNeXt-50}
    \label{fig:xp_resnext_memory}
  \end{subfigure}
  \hfill
  \begin{subfigure}[b]{.245\textwidth}
    \includegraphics[width=1.0\linewidth]{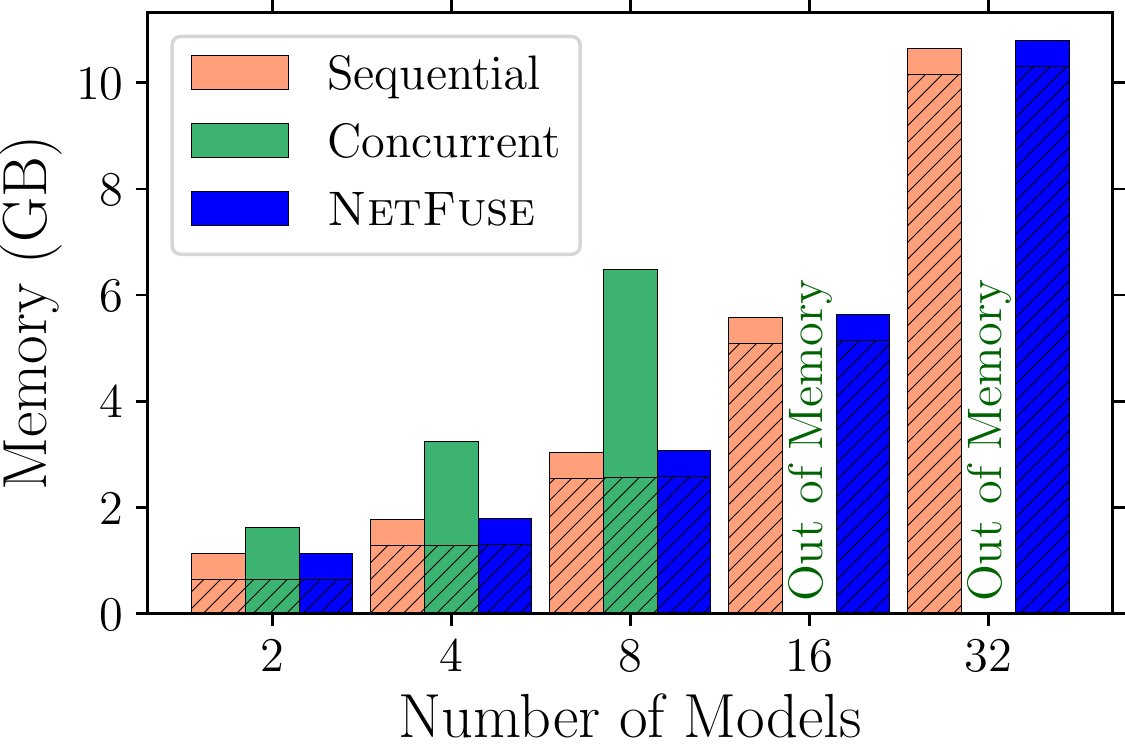}
    \caption{BERT}
    \label{fig:xp_bert_memory}
  \end{subfigure}
  \hfill
  \begin{subfigure}[b]{.245\textwidth}
    \includegraphics[width=1.0\linewidth]{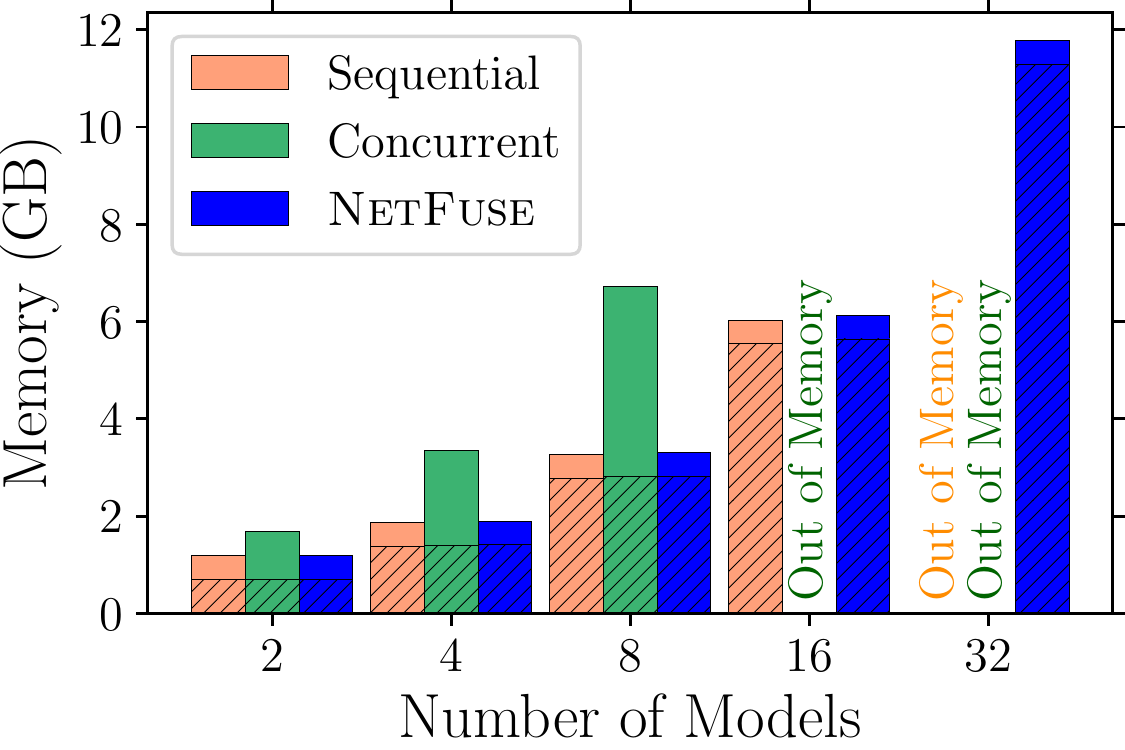}
    \caption{XLNet}
    \label{fig:xp_xlnet_memory}
  \end{subfigure}
  \centering
  \caption{Peak GPU memory usage of \system and the sequential and concurrent baselines for various models on TITAN Xp. The batch size is set to 1. For each vertical bar, the hatched portion denotes the workspace memory, while the solid portion corresponds to the base memory reserved by the framework. TITAN Xp has a total of 12 GB memory.}
  \label{fig:titanxp_memory}
\end{figure*}

The peak GPU memory usage of each configuration is plotted in Figure~\ref{fig:titanxp_memory}.
Compared with the results on V100 (Figure~\ref{fig:v100_memory}), the peak memory usage generally remains unchanged.
However, we have observed some unexpected results. Unlike the results on V100, the concurrent baseline does not run out of memory when running 16 ResNet-50s and ResNeXt-50s.
Moreover, the sequential baseline runs out of memory when merging 32 XLNets.
We have not yet identified the root cause, though we conjecture that PyTorch’s GPU memory caching allocator is exhibiting inconsistent behavior.

\section*{Acknowledgments}
We thank Taebum Kim for his helpful comments on the paper.
This work was supported by Institute for Information \& communications Technology Planning \& Evaluation (IITP) grant funded by the Korea government (MSIT) (No.2015-0-00221, Development of a Unified High-Performance Stack for Diverse Big Data Analytics),
the ICT R\&D program of MSIT/IITP (No.2017-0-01772, Development of QA systems for Video Story Understanding to pass the Video Turing Test),
and Samsung Research Funding \& Incubation Center of Samsung Electronics under project number SRFC-IT2001-03.

\bibliography{netfuse}
\bibliographystyle{acm}


\end{document}